\documentclass{article}
\usepackage{adjustbox}

    \usepackage[final]{neurips_2025}

\usepackage[utf8]{inputenc} 
\usepackage[T1]{fontenc}    
\usepackage{hyperref}       
\usepackage{url}            
\usepackage{booktabs}       

\usepackage{amsfonts}       
\usepackage{nicefrac}       
\usepackage{microtype}      
\usepackage{xcolor}         

\usepackage[labelfont=bf]{caption}
\usepackage{subcaption}
\usepackage{amsmath}
\usepackage{amssymb}
\usepackage{mathtools}
\usepackage{amsthm}
\usepackage{physics}
\usepackage{bm}
\usepackage{siunitx}

\usepackage[capitalize,noabbrev]{cleveref}

\usepackage{comment}

\theoremstyle{plain}
\newtheorem{theorem}{Theorem}[section]
\newtheorem{proposition}[theorem]{Proposition}

\newtheorem{corollary}[theorem]{Corollary}
\theoremstyle{definition}
\newtheorem{definition}[theorem]{Definition}

\theoremstyle{remark}

\definecolor{gray75}{gray}{0.75}

\newcommand{\pr}[1]{\left(#1 \right)} 
\newcommand{\br}[1]{\left[#1 \right]} 
\newcommand{\cbrace}[1]{\left\{#1 \right\}} 

\newcommand{\pd}[2]{\frac{\partial #1}{\partial #2}} 
\newcommand{\diff}{\mathrm{d}}

\renewcommand{\v}[1]{\ensuremath{\mathbf{#1}}} 

\newcommand{\argmin}{\operatornamewithlimits{arg\,min}}

\DeclarePairedDelimiterX{\infdivx}[2]{(}{)}{%
  #1\;\delimsize\|\;#2%
}
\newcommand{\kl}{D_{\text{KL}}\infdivx}

\usepackage{wrapfig,lipsum,booktabs}

\title{Energy Loss Functions for Physical Systems}

\author{%
  Sékou-Oumar~Kaba$^\ast$, Kusha Sareen$^\ast$, Daniel Levy, Siamak~Ravanbakhsh \\
  McGill University\\
  Mila - Quebec Artificial Intelligence Institute\\
}

\begin{document}

\def\thefootnote{$^\ast$}\begin{NoHyper}\footnotetext{Equal contribution. Correspondence: kabaseko@mila.quebec.}\end{NoHyper}

\maketitle

\begin{abstract}
Effectively leveraging prior knowledge of a system’s physics is crucial for applications of machine learning to scientific domains. Previous approaches mostly focused on incorporating physical insights at the architectural level. In this paper, we propose a framework to leverage physical information directly into the loss function for prediction and generative modeling tasks on systems like molecules and spins. We derive \textit{energy loss functions} assuming that each data sample is in thermal equilibrium with respect to an approximate energy landscape. By using the reverse KL divergence with a Boltzmann distribution around the data, we obtain the loss as an energy difference between the data and the model predictions. This perspective also recasts traditional objectives like MSE as energy-based, but with a physically meaningless energy. In contrast, our formulation yields physically grounded loss functions with gradients that better align with valid configurations, while being architecture-agnostic and computationally efficient. The energy loss functions also inherently respect physical symmetries. We demonstrate our approach on molecular generation and spin ground-state prediction and report significant improvements over baselines.
Code is available at \url{https://github.com/kushasareen/energy_loss}.

\end{abstract}

\section{Introduction}

A key challenge in applications of machine learning to the physical sciences is that data can often be scarce and expensive to generate. However, we often have some prior knowledge of the physics of the system of interest, which can be used to design useful inductive biases. A common learning problem involves training a machine learning model to predict configurations of physical systems based on data collected close to equilibrium such as protein folding \citep{noe2020machine,jumper2021highly,abramson2024accurate}, crystal structure prediction \citep{ryan2018crystal,jiao2023crystal,zeni2025generative}, calculation of ground states given Hamiltonian parameters \citep{carrasquilla2017}, or generative modeling of physical systems \citep{gomez2018automatic,sanchez2018inverse}.
A significant body of work has focused on implementing physical inductive biases, such as equivariance at the level of architectures (see e.g. \cite{zhang2023artificial} for a review).

This work explores a complementary direction: embedding physical principles directly into the loss function. The fundamental question we ask is: can loss functions grounded in physical principles provide more effective training signals and yield models that better reflect physically valid configurations compared to generic losses such as the mean-squared error (MSE) and the cross-entropy loss?%

As a response, we propose a framework for deriving \textit{energy loss functions} tailored for physical systems {in the thermal equilibrium regime}. This is motivated by the fact that loss functions can be obtained from a distribution representing the uncertainty around each prediction or data sample. For physical systems {in thermal equilibrium}, the sensible choice is the Boltzmann distribution. Employing the reverse Kullback-Leibler (KL) divergence leads to loss functions that take the form of approximate energy differences between data and predictions. This allows for a more principled quantification of the errors made by the model, which we hypothesize provides better gradients for learning.

Our framework is general in the sense that it encompasses many existing loss functions and allows us to interpret them as energies. 
The energy loss functions also naturally capture relevant symmetries if the underlying energy approximation does. Specifically, they make it so that no loss is incurred by the model for predicting configurations that are related to the data by symmetry. Loss functions that have this property have been suggested for atomistic systems, but they require expensive alignment or minimization procedures \citep{klein2023equivariantflowmatching}, which our framework does not require.
\begin{figure}[t]
     \centering
     \vspace{-7ex}
     \begin{subfigure}[t]{0.3\columnwidth}
         \centering
         \includegraphics[width=1\columnwidth]{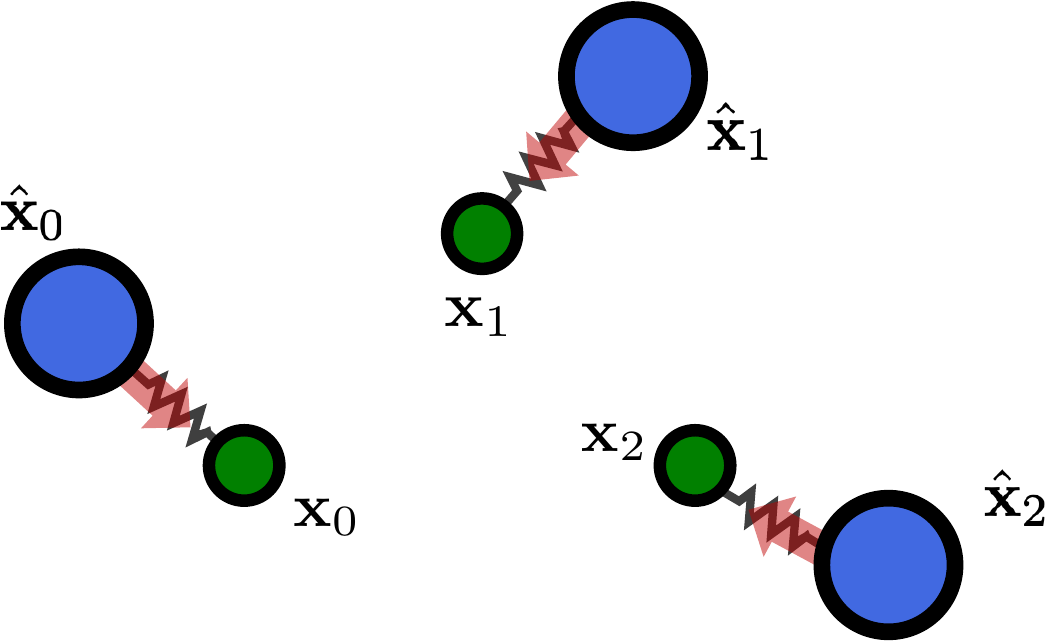}
         \caption{MSE loss}
         \label{subfig:1}
     \end{subfigure}
     \hfill
     \begin{subfigure}[t]{0.23\columnwidth}
         \centering
         \includegraphics[width=1\columnwidth]{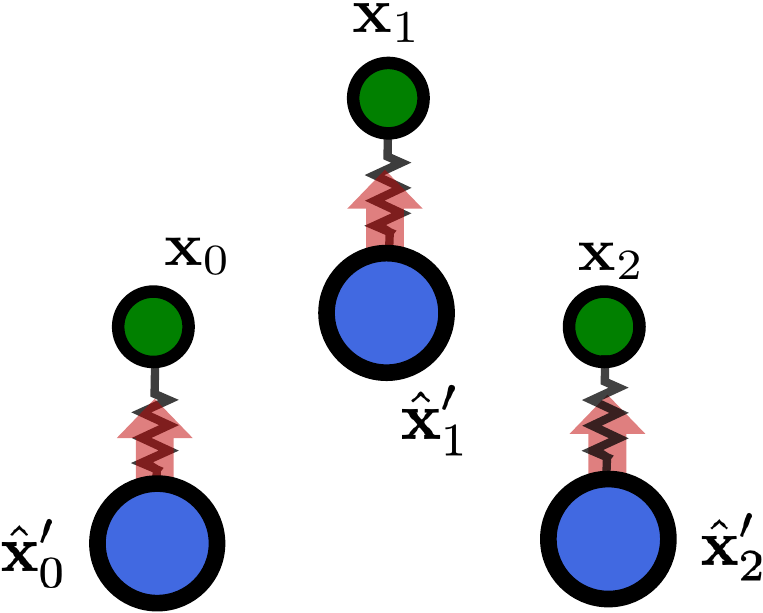}
         \captionsetup{justification=centering}
         \caption{Configuration with same MSE loss as (a)}
         \label{subfig:2}
     \end{subfigure}
     \hfill
     \begin{subfigure}[t]{0.3\columnwidth}
         \centering
         \includegraphics[width=1\columnwidth]{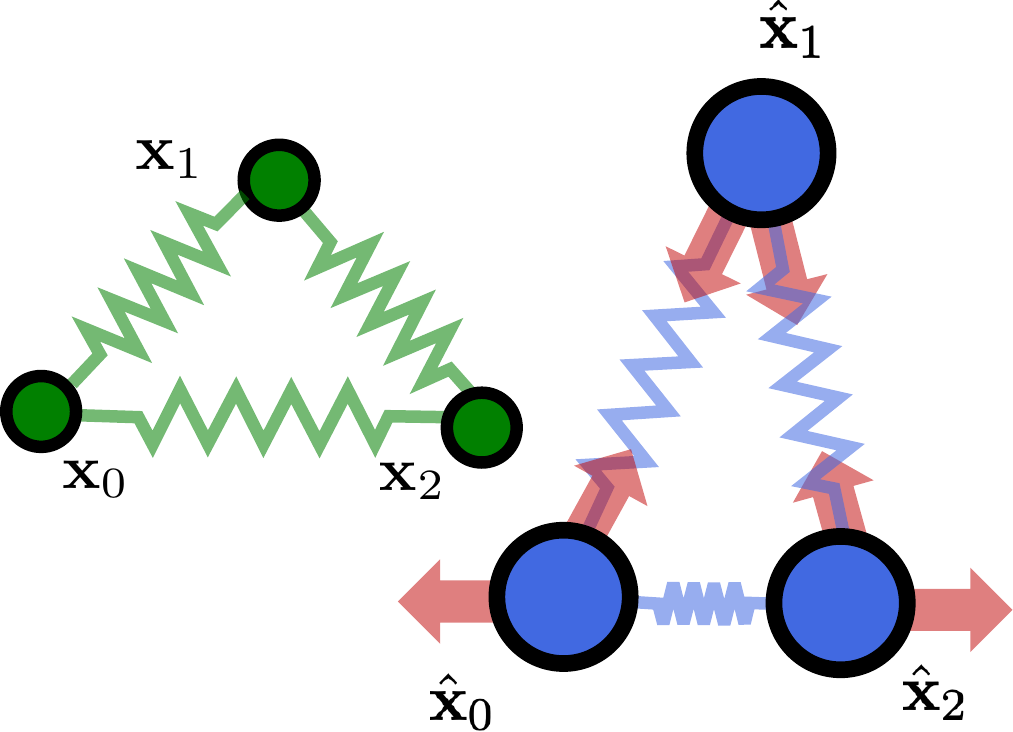}
         \caption{Energy loss}
         \label{subfig:3}
     \end{subfigure}
        \caption{Energy interpretation of loss functions. Ground truth positions are denoted in green and predictions in blue. \textbf{(a)} The MSE loss function for particle positions corresponds to quadratic potential energy centered on the data. \textbf{(b)} This choice is however physically unsound and leads to penalizing the model for configurations that are correct, i.e. related by rigid motion to the target.  \textbf{(c)} A more accurate choice would be to use a loss function based on physically sound energy, which would not suffer from the aforementioned problem. 
        }
        \label{fig:1}
        
\vspace{-2ex}
\end{figure}
This framework finds broad applicability to systems in thermal equilibrium, from direct regression tasks to generative modeling with diffusion models \citep{dickstein2015,ho2020,song2021scorebased}. Note that we consider tasks that can be framed as regression and classification problems with data, not sampling problems where we are given a ground-truth energy (like for example in Boltzmann generators \citep{noe2019boltzmann}).

\textbf{Contributions:} \textbf{(i)} Methodology for deriving loss functions grounded in physical principles by minimizing the reverse KL divergence between a prediction and a Boltzmann distribution centered around data \textbf{(ii)} Instantiation of this framework for atomistic systems that yield distance-based loss functions and an analysis of the invariance properties of these losses \textbf{(iii)} Applications to diffusion models and analysis of the resulting score estimator \textbf{(iv)} Instantiation of this framework for spin systems \textbf{(v)} Empirical evaluation on a range of tasks, showing consistent improvement over baselines.

\vspace{-0.5ex}
\section{Background}
\vspace{-0.25ex}
\subsection{Forward and reverse KL loss functions}
\label{subsec:rkl}

We first consider a regression setting.
Consider the empirical distribution $p_{\mathcal{D}}$ 
associated with the IID dataset $\mathcal{D} = \cbrace{\v{x}^{(i)}, \v{y}^{(i)}}_{i\in \br{N}}$, and a parametric model $f_\theta: \mathbb{R}^d\to \mathbb{R}^k, \v{x}\mapsto \hat{\v{y}}$ associated with the family of conditional distributions $p\pr{\v{y}\mid f_\theta\pr{\v{x}}}$. We take $f_\theta\pr{\v{x}}$ to be the model prediction of the target;  the usual assumption is that conditional distribution is parametrized by a location parameter (like the mean for a Gaussian), and the model is trained to maximize the likelihood of the data $\mathcal{L}\pr{\theta} = -\sum_{i}^N \log p\pr{{\v{y}}^{(i)}\mid f_\theta\pr{\v{x}^{(i)}}}$. The Gaussian assumption for the model results in the Mean Squared Error (MSE) loss function.
For $n$-way classification, the model predicts the logits of a categorical distribution and maximum likelihood yields the cross-entropy loss function.
Maximizing the likelihood is equivalent to minimizing the Kullback-Leibler (KL) divergence. 

In this work, we will consider a reverse KL divergence objective. In the regression case, this amounts to taking the model as deterministic and instead accounting for the uncertainty at the level of the data samples. For regression, we then have $p_{\mathcal{D}}\pr{\v{x}, \v{y}} = \sum_{i}^N \frac{1}{N} \delta\pr{\v{x} - \v{x}_{i}}p\pr{\v{y}\mid \v{y}^{(i)}}$ and $q\pr{\v{x}, \v{y}} = \sum_{i}^N \frac{1}{N} \delta\pr{\v{x} - \v{x}^{(i)}}\delta\pr{\v{y} - f_\theta\pr{\v{x}^{(i)}}}$, where $p\pr{\v{y}\mid \v{y}^{(i)}}$ is a {conditional distribution} that specifies the uncertainty around each target. The reverse KL objective is then
\begin{align}\label{eq:rkl}
\kl{q}{p_{\mathcal{D}}} & = \mathbb{E}_{q}\br{\log q\pr{\v{x}, \v{y}} - \log p_{\mathcal{D}}\pr{\v{x}, \v{y}}}
= - \sum_{i}^N \log p\pr{f_\theta\pr{\v{x}^{(i)}}\mid {\v{y}}^{(i)}}
\end{align}
For classification, the model distribution is still a categorical distribution parametrized by logits to ensure differentiability, but the distribution associated with data samples can be general.

In general, the reverse KL divergence is not equal to the forward KL divergence. Instead, it gives the likelihood of the prediction given an uncertainty model for targets. However, it is exactly equal when the sample point and the parameter can be swapped in the distribution $p$. It is for example the case when $p$ is chosen as Gaussian. Our general goal will be to define more appropriate distributions $p(\hat{\v{y}}\mid \v{y})$ for the loss function. As we will see, the reverse KL formulation is convenient since it enables defining these distributions only around each data sample.

\vspace{-0.25ex}
\subsection{Diffusion models}
\label{subsec:generative}
We also consider generative modeling with diffusion models as another use case for more informed energy loss functions.
This class of generative models has proven powerful, as they can efficiently learn interpolations between a prior distribution and the data distribution \citep{albergo2023stochastic}.
The objective is typically formulated as a noise prediction task \citep{ho2020}
\begin{align}
    \mathcal{J}\pr{\theta} = \int_0^1 \mathbb{E}_{{\v{x}\sim p(\v{x}), \bm{\epsilon}_t\sim p\pr{\bm{\epsilon}_t}}} \br{ w_t \norm{\bm{\epsilon} - \hat{\bm{\epsilon}}_\theta}^2} \diff t
\end{align}
where the noise prediction is the output of a neural network $\hat{\bm{\epsilon}}_\theta = f_{\theta}\pr{\v{x}_t, t}$, $\v{x}_t = \alpha_t \v{x} + \sigma_t \bm{\epsilon}$, $\sigma_t, \alpha_t$ define the noise schedule and $w_t$ is a weighting factor. In practice, the expectation is estimated by Monte Carlo. The objective also admits an interpretation as {denoising score matching} \citep{vincent2011connection}, with the optimal noise prediction satisfying $\epsilon^*\pr{\v{x}_t, t} = -{\sigma_t} \nabla_{\v{x}_t}\log p\pr{\v{x}_t}$.

The loss can be equivalently seen as prediction of the data sample, with appropriate reweighting, see e.g. \citet{kingma2024understanding}. With the sample prediction defined as $\hat{\v{x}}_{\theta} \equiv \frac{\pr{\v{x}_t - \sigma_t \hat{\bm{\epsilon}}_{\theta}\pr{\v{x}_t, t}}}{\alpha_t}$, we have:
\begin{align} \label{eq:pred_loss}
    \mathcal{J}\pr{\theta} = \int_0^1 \mathbb{E}_{{\v{x}\sim p(\v{x}), \bm{\epsilon}_t\sim p\pr{\bm{\epsilon}_t}}}\br{\frac{w_t \alpha_t^2}{\sigma_t^2} \norm{\v{x} - \hat{\v{x}}_\theta}^2} \diff t
\end{align}
 yielding a regression-type objective with the MSE loss.

\vspace{-0.5ex}
\section{Energy Loss Functions}
\label{sec:energy_loss}
In \cref{subsec:rkl}, we saw that loss functions can be obtained through a reverse KL formulation with respect to a conditional distribution $p\pr{\hat{\v{y}} \mid \v{y}}$ centered on the data. Importantly, the conditional distribution $p$ is always an uncertainty model; as such, there is not necessarily a \textit{true} one. 

We will define the conditional distribution $p$ as a Boltzmann distribution
\begin{align}\label{eq:boltzmann}
p(\hat{\v{y}} \mid \v{y}) = \frac{\exp\pr{- E(\hat{\v{y}}, \v{y}) / T}}{Z(\v{y}, T)}
\end{align}
where $E: \mathbb{R}^k \times \mathbb{R}^k \to \mathbb{R}$ is related to the \textit{physical} potential energy of the system around the data point $\v{y}$, $T$ is the temperature and $Z(\v{y}, T)$ is the partition function. We assume the system is observed in physically likely configurations; hence, each data point $\v{y}$ is modeled as an approximate local minimum in the energy landscape.
The use of Boltzmann distributions to model the uncertainty around such configurations is natural and can be motivated from first principles \citep{jaynes1957information,pathria2017statistical}. It is the steady-state distribution of a system undergoing stochastic dynamics in contact with a reservoir at temperature $T$ (see derivation in \cref{apd:boltzmann}).

Assuming a general Boltzmann distribution, the reverse KL divergence \cref{eq:rkl} loss function we obtain for the continuous case is
\begin{align}
 \mathcal{J}\pr{\theta} &= - \sum_i^N \log p\pr{\hat{\v{y}}^{(i)}_\theta\mid {\v{y}}^{(i)}} = \sum_i^N \frac{E(\hat{\v{y}}^{(i)}_\theta, {\v{y}}^{(i)})}{T} + \log Z({\v{y}}^{(i)}, T),\label{eq:ploss}
\end{align}
where the log-partition function does not depend on the parameters.
The model is penalized for errors by an amount given by the approximate increase in energy with respect to the data sample. 

This picture allows for obtaining a physical interpretation of different conditional distributions and loss functions depending on the choice of energy $E(\hat{\v{y}}, \v{y})$. The Gaussian conditional distribution is obtained with $T=2\sigma^2$ and isotropic harmonic potential energy centered around $\v{y}$:
\begin{align}\label{eq:harmonic}
E(\hat{\v{y}}, \v{y}) = \norm{\hat{\v{y}} - \v{y}}^2.
\end{align}

We can now justify our choice of the reverse KL estimation over maximum likelihood estimation. First, in the reverse KL case, the partition function, which could be challenging to evaluate for some energies, does not depend on the model parameters $\theta$. Second, we only need to define potential functions around each data sample ${\v{y}}^{(i)}$, rather than around each prediction. This is a significant advantage, as we can expect some predictions to be poor, leading to configurations that are not approximate equilibria and to nonsensical energies.

\vspace{-0.25ex}
\subsection{Desiderata for energy functions}

There is considerable freedom in the choice of the energy function. One fundamental criterion is agreement with the system's underlying physics, but this is not the only one. An appropriate energy should, in addition, satisfy the following desiderata:
\begin{enumerate}
\item \textbf{Minimized at the data and symmetries:} The minimizer of the energy function $E(\hat{\v{y}}, \v{y})$ should be the data $\v{y}$ (and its symmetry equivalents). Many tasks require regressing to the data even if it is not the minimum of the true energy landscape.
\item \textbf{Optimization stability:} The gradient of the energy function $\nabla_{\hat{\v{y}}} E(\hat{\v{y}}, \v{y})$ should be smooth and bounded to ensure well-behaved optimization with gradient-based methods.
\item \textbf{Fast evaluation:} Evaluation of the energy and its derivative should be efficient and compatible with automatic differentiation. 
\end{enumerate}

Based on this, we argue that one \textit{should not} often use the true energy function, even if it is known, since it may violate all the criteria. The energy landscapes of systems of interest typically admit multiple local minima and are highly rugged \citep{mezard1987spin,frauenfelder1991energy,wales2000energy}. The cost of evaluating the energy can also be prohibitive \citep{schuch2009computational}.

\vspace{-0.5ex}
\section{Energies for Atomistic Systems}

We consider energies associated with the positions of $n$ atoms in $d$ dimensions, such that $\hat{\v{y}},\v{y}\in \mathbb{R}^{n\times d}$. The potential energy \cref{eq:harmonic} leading to the Gaussian distribution is poorly motivated from the physical point of view. It describes the effect of an external force bringing back particles to position $\v{y}$. However, a realistic potential energy should model \textit{interactions} between particles (see \cref{subfig:3}).

Many approximations exist for the potential energies of physical systems around equilibrium. For atomic systems, the Morse potential \citep{morse1929diatomic} and the Lennard-Jones potential \citep{lennard-jones} are examples of popular models. However, using these potentials for the loss \cref{eq:ploss} can pose challenges for optimization, as they have highly nonlinear gradients that can explode or vanish. A simple approximation that avoids this issue and that is much more principled than the MSE potential is to use a quadratic pair potential of the form
\begin{align}\label{eq:loss}
E(\hat{\v{y}}, \v{y}) = \sum_{i,j}^n \frac{1}{2} k_{ij}\pr{\v{y}} \left(\lVert\v{y}_i -\v{y}_j \rVert - \lVert\hat{\v{y}}_i -\hat{\v{y}}_j \rVert \right)^2
\end{align}
where the indices $i,j$ are taken over particles. This is the general form of a second-order Taylor approximation in pairwise distances of an interaction potential (see \cref{apd:taylor}). Motivated by the fact that coordinate regression can lead to poor realism due to inconsistencies with bond lengths, this type of distance-dependent loss function has been used heuristically as a regularizer in some applications \citep{peng2023moldiff,yang2023chemically,abramson2024accurate}, but to the best of our knowledge, not as a primary objective. Note that this is different from directly predicting the distances \citep{simm2019generative,nesterov20203dmolnet,xu2021learning}.

There is significant freedom in the choice of the coefficients $k_{ij}\pr{\v{y}}$. We propose simple heuristics to set these coefficients.
First, we consider setting the coefficients are set to a constant value $k_{ij}\pr{\v{y}} = k$. Note that this can be obtained from Taylor approximation of the Morse potential (see \cref{apd:taylor}).
Second, we consider setting the coefficient as a decreasing function of the distance between two atoms $k_{ij}\pr{\v{y}} = f(\norm{\v{y}_i - \v{y}_j})$ to capture the fact that interactions between particles decrease at long range. We consider inverse, inverse-squared, and exponential decay dependence of $f$ on the distance. Taylor approximation of the Lennard-Jones potential yields inverse squared distance dependence (see \cref{apd:taylor}). Other possibilities can be considered: in general, given an interaction potential between particles, the coefficients can be obtained by a second-order Taylor expansion.

\vspace{-0.25ex}
\subsection{Invariance properties}
\label{subsec:invariance}
An important property of energy loss functions is that they respect the symmetries of the associated physical energy. We formalize this in the following way:
\begin{definition}[Invariant loss function]
A loss function $l: \mathbb{R}^k \times \mathbb{R}^k \to \mathbb{R}$ between a prediction and a target is invariant to the action of the group $G$ on $\mathbb{R}^k$ if
\begin{align}
    l\pr{g\cdot \hat{\v{y}}, \v{y}} = l\pr{\hat{\v{y}}, g\cdot \v{y}} = l\pr{\hat{\v{y}}, \v{y}}, \quad \forall g\in G, \hat{\v{y}}, \v{y} \in \mathbb{R}^k
\end{align} 
\end{definition}
An invariant loss function essentially compares input and targets {up to} transformations in $G$.
{An example of a common $SE(3)$-invariant loss function is to apply the Kabsch algorithm \citep{kabsch1976solution} to find an optimal alignment between a predicted structure and a target, and to use the MSE after applying the alignment \citep{klein2023equivariantflowmatching}.}
It has been shown that in cases where there are multiple possible symmetry-related predictions for a given input-- so-called symmetry-breaking predictions \citep{smidt2021finding,kaba2023symmetry}-- non-invariant loss functions exhibit pathological behaviour \citep{xie2024equivariant,jing2024alphafold,lawrence2025improving}. For example, the MSE is minimized when the prediction is the mean of the possible targets rather than for any of them. There are multiple ways to define these invariant losses, which are analogous to the different ways in which invariant neural networks can be designed (see \cref{apd:invariant} for more discussion).

\begin{figure}[t]
     \centering
         \vspace{-7ex}
     \begin{subfigure}[t]{0.22\columnwidth}
         \centering
         \includegraphics[width=1\columnwidth]{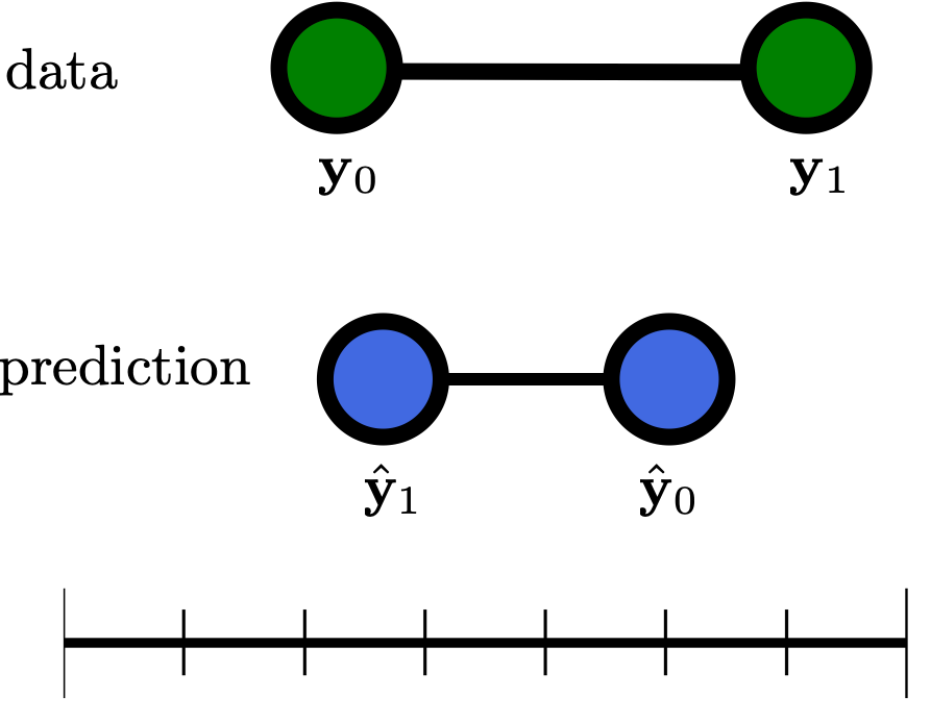}
     \end{subfigure}
     \hfill
     \begin{subfigure}[t]{0.32\columnwidth}
         \centering
         \includegraphics[width=1\columnwidth]{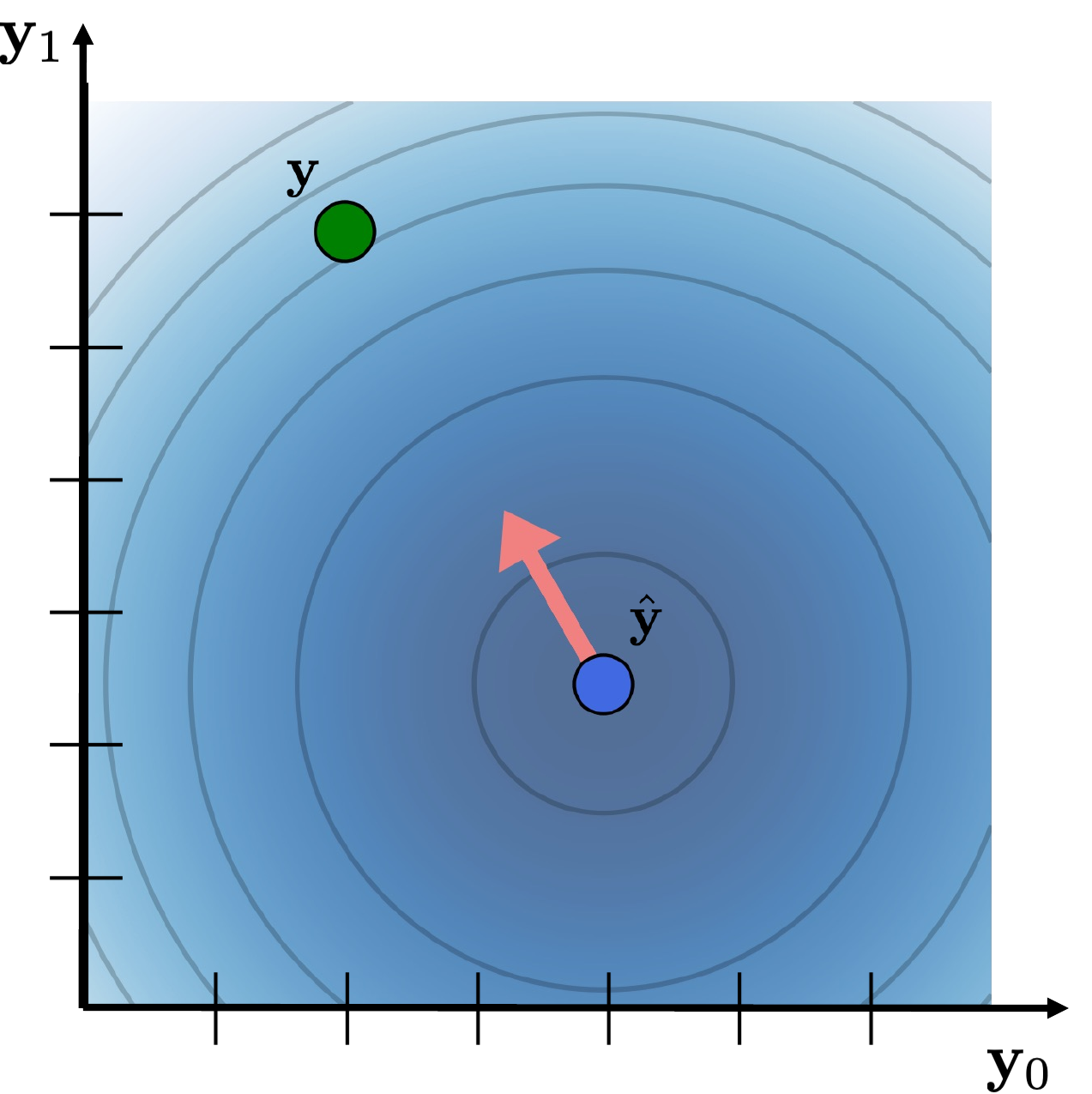}
         \caption{MSE loss}
         \label{subfig:2.2}
     \end{subfigure}
     \hfill
     \begin{subfigure}[t]{0.32\columnwidth}
         \centering
         \includegraphics[width=1\columnwidth]{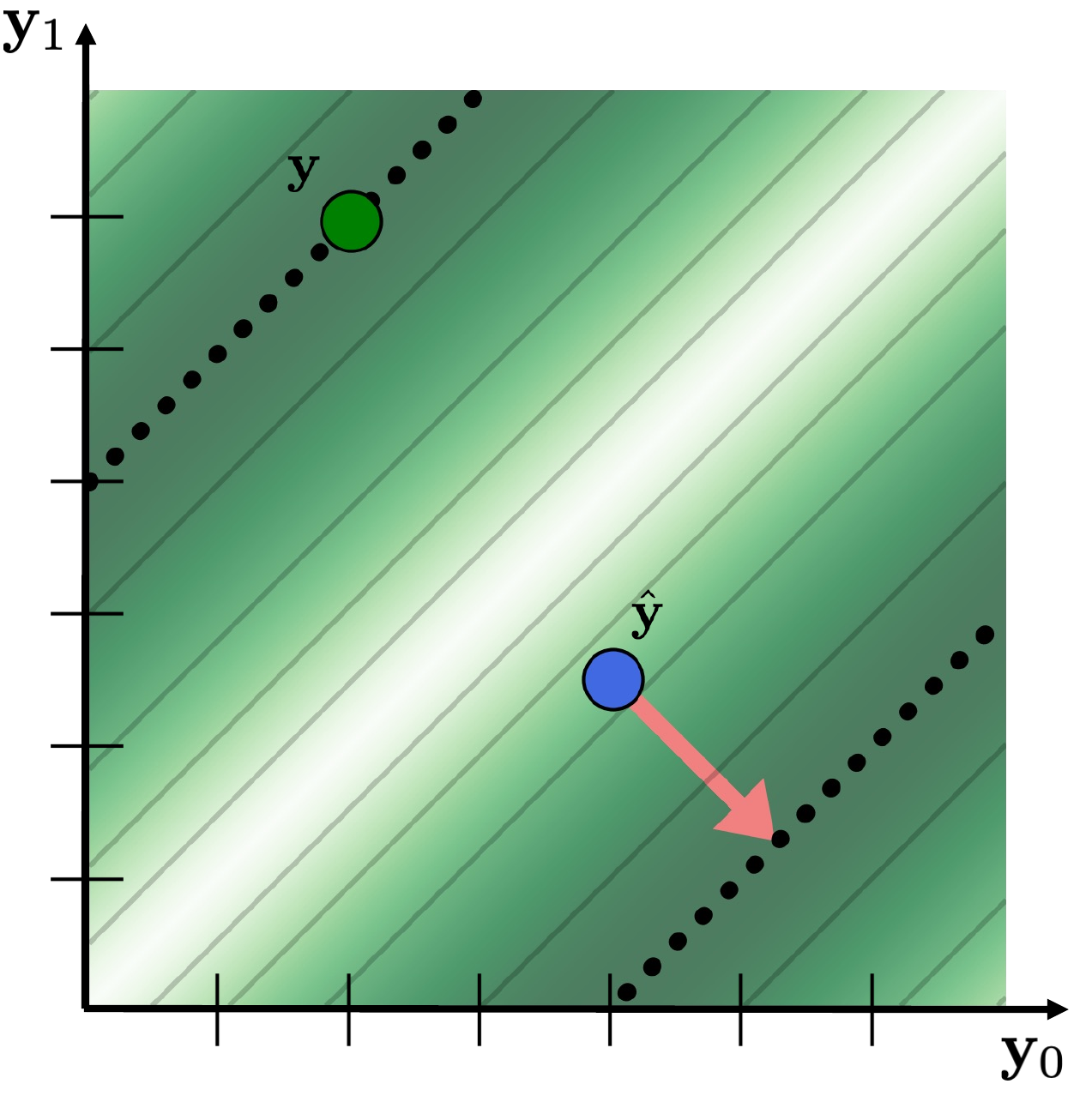}
         \caption{Energy loss}
         \label{subfig:2.3}
     \end{subfigure}
        \caption{Loss landscapes. The model has to predict the positions of two particles in one dimension. The prediction for the first particle $\hat{\v{y}}_0$ is closer to the ground-truth for the second particle $\v{y}_1$ and vice-versa. \textbf{(a)} The MSE minimizes the forward KL divergence between a Gaussian model distribution (blue) and the data distribution (green). It does not capture the symmetry. \textbf{(b)} The energy loss is obtained via the reverse KL with the pair energy and admits a family of minimizers associated with symmetries. It results in a gradient that points towards the closest correct configuration.}
        \label{fig:loss}
    \vspace{-2ex}
\end{figure}

It is easy to see that the energy loss \cref{eq:ploss} is invariant to $G=E(d)$ if $k_{ij}\pr{\v{y}}$ is invariant, since it then only depends on invariant distances.
This is analogous to how invariant functions can be built from scalars \cite{villar2021scalars}.
These, however, are not the only symmetries of the loss. The energy loss function is additionally invariant to permutations that correspond to the symmetries of the ground-truth distance matrix. Denote the distance matrix of the data by $\Delta y_{ij} = \lVert\v{y}_i -\v{y}_j \rVert$ and the automorphism group of a matrix $m\in \mathbb{R}^{n\times n}$ as $\mathrm{Aut}\pr{m} \subseteq S_n$ where the automorphisms act on the matrix by conjugation. We then have the following:
\begin{proposition}
\label{prop:symmetry}
The loss function \cref{eq:loss} is invariant to the group
\begin{align}
G = E(d) \times (\mathrm{Aut}\pr{k\pr{\v{y}}}\cap \mathrm{Aut}\pr{\Delta y}).
\end{align}
\end{proposition}
All the proofs follow in \cref{{apd:proofs}}.
This allows us to characterize the family of loss minimizers:
\begin{corollary}
\label{corol:min}
For any ${\v{y}}\in \mathbb{R}^{n\times d}$ and $k_{ij}\pr{\v{y}} > 0$,
\begin{align}
    \argmin_{\hat{\v{y}}\in \mathbb{R}^{n\times d}} E(\hat{\v{y}}, \v{y}) = \cbrace{g\cdot \v{y} \ \mid \ g\in G}.
\end{align}
\end{corollary}
The loss landscape, therefore, presents a family of global minimizers associated with symmetries. 
We hypothesize that the symmetry is beneficial for learning, since it allows the model to regress to any target that is equivalent to the data by symmetry, as shown in the example in \cref{fig:loss}.
As our experimental results show, the benefits of invariance in the loss function are different and complementary to that of equivariance of the architecture. Equivariance guarantees that the output changes predictably under transformations of the input. However, it does not guarantee that the correct output will be learned. Invariant loss functions make the learning task easier by allowing to regress to any symmetry related configurations, which equivariance with a non-invariant loss does not allow.

\vspace{-0.25ex}
\subsection{Diffusion models with distance-based loss functions}
\label{subsec:difusion_distance_loss}
The training objective of diffusion models involves the prediction of a data sample from a noisy latent one. We suggest that the energy loss functions can be used as a straightforward replacement for the MSE in these objectives. Similar distance-based objectives have been previously used in the context of diffusion models \citep{yang2023chemically, abramson2024accurate, cognolato2025d4}. However, it is not immediately clear that using such objectives results in learning correct score estimates. Here, we show that this is indeed the case under some conditions.

Consider the energy loss function \cref{eq:loss} with constant coefficients $k_{ij}\pr{\v{x}} = k$. The loss function computes the MSE between distance matrices for the data and the sample prediction, given by $\Delta x_{ij} = d\pr{\v{x}} = \norm{\v{x}_i- \v{x}_j}$. Denote the Jacobian of this function by $J(\v{x}) = \nabla_{\v{x}} d\pr{\v{x}}\in \mathbb{R}^{dn\times n^2}$. We will seek to characterize the minimizers of a diffusion model trained using this loss,
\begin{align}
\hat{\bm{\epsilon}}^* \in \argmin_{\hat{\bm{\epsilon}}\in \mathbb{R}^{n\times d}} \ \mathbb{E}_{{\v{x}\sim p(\v{x}_0), \bm{\epsilon}_t\sim p\pr{\bm{\epsilon}_t}}}\br{ E\pr{{\frac{\v{x}_t - \sigma_t \hat{\bm{\epsilon}}\pr{\v{x}_t, t}}{\alpha_t}}, \v{x}}}
\end{align}
The following result allows us to obtain an approximation of this set, valid for small noise scales:
\begin{proposition}
\label{prop:score}
Let $p\pr{\v{x}_t}$ be a continuously differentiable, $SE(d)$-invariant density. Assume $\abs{\hat{\bm{\epsilon}}}$ is bounded. For small $\sigma_t$,
\begin{align}
\hat{\bm{\epsilon}}^* \approx - {\sigma_t} \nabla_{\v{x}_t} \log p\pr{\v{x}_t} + \v{v}, \quad  \v{v} \in \text{ker}\pr{J\pr{\v{x}_t}}
\end{align}
In addition, the minimum norm minimizer $\hat{\bm{\epsilon}}^*_{\text{dist}}$ is given by
\begin{align}
\hat{\bm{\epsilon}}^*_{\text{dist}} 
\approx - {\sigma_t} \nabla_{\v{x}_t}\log p\pr{\v{x}_t}
\end{align}
\end{proposition}
The set of minimizers is therefore given by the true score, up to a translation in the direction of rigid motions. The second fact follows since for an invariant measure, the score is orthogonal to the Lie algebra generators.
We can also show that due to its invariance, the distance-based loss function offers a reduction in variance with respect to the MSE:
\begin{proposition}
\label{prop:variance}
Let $p\pr{\v{x}_t}$ be a continuously differentiable, $SE(d)$-invariant density. Denote by $\hat{\bm{\epsilon}}^*_{\text{dist}}$ and $\hat{\bm{\epsilon}}^*_{\text{MSE}}$ the minimum norm minimizers of the Monte-Carlo estimators of the energy loss and MSE loss,  respectively. For small $\sigma_t$,
\begin{align}
&\text{Bias}\br{\hat{\bm{\epsilon}}^*_{\text{dist}}} \approx 0, &&\text{Var}\br{\hat{\bm{\epsilon}}^*_{\text{dist}}} \lesssim \text{Var}\br{\hat{\bm{\epsilon}}^*_{\text{MSE}}}
\end{align}
\end{proposition}
Note that surprisingly, even though these results are in principle only valid for the energy loss function with constant coefficient $k_{ij}\pr{\v{x}}$, in our results of \cref{subsec:molecule}, the variants using more physically motivated coefficients still performed better empirically. We hypothesize that this is because the evaluation of the model assesses the physical plausibility of the samples rather than the agreement between the learned and data distributions. Incorporating physical information in the loss function can therefore be beneficial, even though (or because) it biases the score estimate. 

\vspace{-0.25ex}
\subsection{Linear scaling and rigidity theory}
\label{subsec:rigidity}
One potential downside of using the energy loss of \cref{eq:loss} is that it has a quadratic number of terms in the number of particles $N$, in contrast to the linear number of terms in typical losses such as MSE loss. While in many architectures-- such as transformers or densely connected graph neural networks-- the quadratic cost of operations in the network makes this a non-issue, the feasibility of linear scaling may prove valuable for applications involving a large number of particles, e.g., modelling macromolecules or crystals with large unit cells. 

Fortunately, a solution is provided by rigidity theory.
Results in rigidity theory \citep{laman1970graphs, asimow1978rigidity} provide the conditions for recovering the coordinates of a point cloud from a \textit{linear} number of pairwise distances. In this work, we consider a construction for sparse rigid graphs, reducing the computational cost of energy loss without affecting its global optima (see \cref{apd:rigidity} for more background and \cref{apd:sparse_timing} for wall-times of different loss calculations).

\vspace{-0.5ex}
\section{Energy Loss for Discrete Systems}
The energy loss formulation can be leveraged for other types of systems. We derive a version for the discrete case, which can replace the cross-entropy loss function. Denote logits predictions as $\v{z}^{(i)}_{\theta,j}$ and the associated categorical distribution as $q\pr{{\hat{\v{y}}}\mid \v{z}^{(i)}_{\theta}}$. The reverse KL between the model distribution and a Boltzmann distribution around the data \cref{eq:boltzmann} is given by
\begin{align}
\label{eq:discrete}
 \mathcal{J}\pr{\theta} &= \frac{1}{T}\sum_i^N \br{
  {\mathbb{E}_{q\pr{{\hat{\v{y}}}\mid \v{z}^{(i)}_{\theta}}}\br{E\pr{\hat{\v{y}}, \v{y}^{(i)}}} - T S\br{q\pr{{\hat{\v{y}}}\mid \v{z}^{(i)}_{\theta}}} + T \log Z\pr{\v{y}^{(i)}, T}}
 }
\end{align}
where $\v{z}^{(i)}_{\theta,j}$ is the model prediction for the logits associated with class $j$ and $S[q]$ is the entropy of $q$. The loss is therefore proportional to the free energy difference.
The last term is the negative free energy at the data, and does not depend on the parameters. The loss function, therefore, simply reduces to the variational free energy of the prediction.

\vspace{-0.25ex}
\subsection{Application to spin systems}

We consider modeling systems of spins as an application of the discrete formulation. Predicting configurations of these systems with machine learning models is a problem of high interest in physics \citep{carrasquilla2017,pahng2020predicting} and in combinatorial optimization \citep{fu1986}.
We will be interested specifically in systems on a square lattice $\Lambda$ such that $\hat{\v{y}},\v{y}\in \cbrace{1,-1}^{\Lambda}$. We consider Ising-type Hamiltonians of the form $E\pr{\v{y}} = -\frac{1}{2} \sum_{ij}^\Lambda J_{ij} \v{y}_i\v{y}_j$
where the coupling $-1 \leq J_{ij} \leq 1$ is non-zero only for neighboring sites in the lattice $\Lambda$, but does not necessarily exhibit any symmetry. Systems with unstructured couplings are known as spin glasses \citep{mezard1987spin} and often exhibit a large number of local energy minima.

Energy loss functions of the form \cref{eq:discrete} can be used for classification of spin configurations. We suggest to use an approximate local energy around the data defined as
\begin{align}
\label{eq:spinloss}
E\pr{\hat{\v{y}},\v{y}}= \sum_{i}^\Lambda  h^{\text{LF}}_i\pr{\v{y}}\hat{\v{y}}_i
\end{align}
where the local field is given by $h^{\text{LF}}_i\pr{\v{y}}= \sum_{j}^\Lambda (J_{ij} + h^0) \v{y}_j$.
The local field energy captures the change in energy from flipping a spin in the configuration $\v{y}$. It therefore provides an appropriate way to quantify deviations from that configuration: large values of local field are associated with spins that result in large increases of energy and that should be weighted more importantly in the loss.

An alternative would be to use the true energy instead. The objective \cref{eq:discrete} would then be interpreted as entropy-regularized energy minimization.
This would be expected to perform well in strict terms of minimizing the energy. However, the true energy is not a classification objective, since it does not make use of the data. In addition, it can exhibit a large number of local minima. By contrast, the local energy loss \cref{eq:spinloss} is convex, due to the linear dependence in $\hat{\v{y}}$. It also admits the data point $\v{y}$ as its unique minimum if $h^0> 4$ (see \cref{apd:spin}). If the data is a ground state of the true energy, $h^0> 0$ is sufficient. The energy loss is therefore a proper classification objective.

\vspace{-0.5ex}
\section{Experiments}
\label{sec:experiments}
\vspace{-0.25ex}
\subsection{Regular shape prediction}
\begin{figure}[t]
    \centering
    \vspace{-6ex}
    \includegraphics[width=1\linewidth]{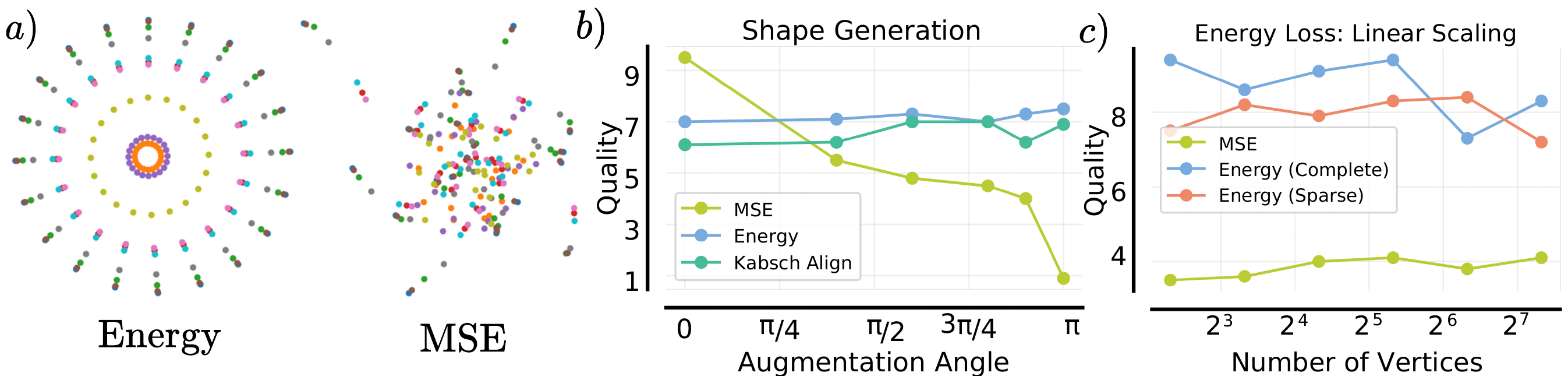}
    \caption{Regular shape prediction results. \textbf{(a)} Typical samples from optimal models trained with MSE and energy loss when $\theta_{aug} = \pi$. \textbf{(b)} The impact of $\theta_{aug}$ on sample quality. We can see as $\theta_{aug}$ increases, MSE performance deteriorates but the invariant losses (Energy and Kabsch Align) remain performant. \textbf{(c)} As the number of shape vertices scales, a sparse version of the energy loss remains equally performant as a complete-edge energy loss using only $O(N)$ operations.}
    \label{fig:shape_generation}
    \vspace{-2ex}
\end{figure}
\paragraph{Experiment Setup.}
To develop an understanding of energy loss functions, we propose a simple task where the goal is to generate regular shapes in two dimensions. Given a radius, a model is tasked with predicting the $N$ vertices of a regular polygon of that radius. The dataset is constructed by sampling regular polygons of a radius $r \sim U[0.3, 5]$ and then applying an augmentation by randomly rotating the shape by an angle in $U[-\theta_{\text{aug}}, \theta_{\text{aug}}]$. Prediction is performed using two hidden-layer MLP. We compare standard MSE loss with the atomic energy loss using exponential coefficients, an SE(2)-invariant loss using the Kabsch algorithm to align points, and a version of the energy loss using sparse rigid graphs. We empirically confirm that the sparse graphs are globally rigid w.h.p. in \cref{apd:rigidity}.
To evaluate, we introduce a quality metric based on the regularity of the angular differences and the radial variation in a given shape. 
Intuitively, for a regular shape the angular difference variation $\sigma_{\Delta_{angle}}$ and the radial variation $\sigma_{radius}$ across points should be small. A full definition follows in \cref{apd:shape}.

\textbf{Results.} \autoref{fig:shape_generation} shows that the energy loss and other invariant losses continue to produce high-quality shapes when rotation augmentation is applied whereas MSE fails. Additionally, the sparse energy loss maintains nearly the same performance as the number of vertices $N$ increases, while reducing computation by $O(N)$ operations. Interestingly, models trained with an invariant loss automatically learn to produce canonical orientations of shapes.  

\vspace{-0.25ex}
\subsection{Molecule generation}
\begin{wraptable}{r}{9.2cm}
\vspace{-6ex}
\footnotesize
\caption{Metrics for GDM-aug on GEOM-Drugs.}\label{tab:geom_drugs}
\begin{tabular}{@{}lcccc@{}}
  \toprule
  Loss & Mol. stab. (\%) & Atom stab. (\%) & Valid. (\%) & Unique (\%) \\
  \midrule
    MSE    & 0.8  & 85.6 & \textbf{94.8} & 100 \\
    Energy & \textbf{24.6} & \textbf{96.0} & 89.7 & 100 \\
  \bottomrule
\end{tabular}
\end{wraptable}

\paragraph{Experiment Setup.}
First, we train diffusion models to unconditionally generate molecules in the QM9 dataset \citep{Ramakrishnan2014}.  We evaluate the performance of the energy loss when training EGNN diffusion models (EDM) \citep{hoogeboom2022equivariantdiffusionmoleculegeneration}, GNN diffusion models with and without data augmentation (GDM and GDM-aug) and near state-of-the-art joint 2D \& 3D diffusion models (JODO) \cite{huang2023learningjoint2d}. As baselines, we compare the convergence properties to models trained with MSE and a Kabsch-aligned MSE \citep{kabsch1976solution}. Exponential coefficients are chosen for the energy loss. 
Additionally, we compare with a version of the energy loss using sparse rigid graphs. 

\begin{table}[h!]
\footnotesize 
\vspace{-2ex}
  \caption{{Evaluation metrics for GDM-aug on QM9.}} 
  \label{tab:qm9}
  \centering
  \begin{tabular}{lcccc}
    \toprule
    Loss & Molecule stability (\%) & Atom stablity (\%) & Validity (\%) & Uniqueness (\%)\\
    \midrule
    \multicolumn{5}{l}{\textbf{GDM-aug}} \\
    \midrule
    MSE & 83.7 $\pm$ 2.3 & 98.3 $\pm$ 0.004 & 93.6 $\pm$ 1.7 & 100.0 $\pm$ 0.0 \\
    Kabsch align & 82.3 $\pm$ 0.5 & 97.8 $\pm$ 0.004 & 90.8 $\pm$ 2.0 & 100.0 $\pm$ 0.0 \\
    Energy & \textbf{89.8} $\pm$ 2.8 & \textbf{99.3} $\pm$ 0.3 & \textbf{97.7} $\pm$ 1.4 & 99.9 $\pm$ 0.002 \\ 
    Energy (sparse) & 89.1 $\pm$ 0.9 & 99.0 $\pm$ 0.1 & 97.4 $\pm$ 2.5 & 100 $\pm$ 0.0 \\
    \midrule
    \multicolumn{5}{l}{\textbf{EDM}} \\
    \midrule
MSE & 82.4 $\pm$ 3.4 & 98.8 $\pm$ 1.7 & 93.0 $\pm$ 2.5 & 99.89 $\pm$ 0.32 \\
    \bottomrule
  \end{tabular}
\end{table}

We also generate large molecules with GDM and GDM-aug using the GEOM-Drugs dataset \citep{axelrod2022geomenergyannotatedmolecularconformations}, comparing the MSE and energy loss. 
A similar evaluation setup to \citep{satorras2022enequivariantnormalizingflows,hoogeboom2022equivariantdiffusionmoleculegeneration} is used for GDM and EDM while JODO uses a broader set of 3D and align metrics \cite{huang2023learningjoint2d}. Since the MSE is no longer the optimization objective, we no longer report the ELBO, which depends on the MSE. Instead, we evaluate the method using several desirable features of generated molecules relevant to the drug discovery pipeline: atom stability, molecule stability, validity, and uniqueness. Additionally, for JODO, we compute the Maximum Mean Discrepancy (MMD) for bond lengths, bond angles and dihedral angles against the data distribution as well as the Fréchet ChemNet Distance (FCD) \cite{fcd2018}.
For all settings, we conduct exhaustive sweeps for learning rate and the weighting between the loss on positions and atom types. All comparisons are compute-matched.

\begin{figure}[t!]
    \centering
    \includegraphics[width=1\linewidth]{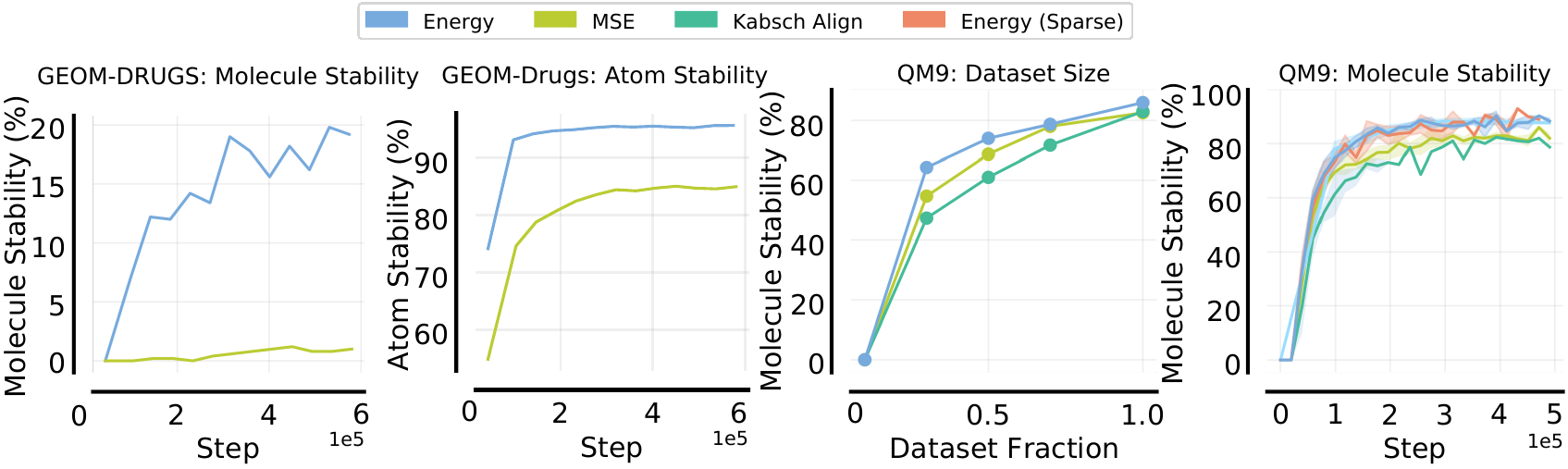}
    \caption{Molecule generation results. \textbf{(Left)} We observe a dramatic improvement on stability metrics for the GEOM-Drugs dataset, demonstrating the scalability of our approach. \textbf{(Right)} On QM9, energy loss improves metrics over all baselines. This is especially present in the low data regime where energy loss gives +$10\%$ molecule stability over MSE.}
    \label{fig:molecule_generation}
    \vspace{-2ex}
\end{figure}

\paragraph{Results.} \autoref{fig:molecule_generation} shows the energy loss results in faster convergence and better optima over baselines\footnote{We note that our results with MSE are better than those reported in \citet{hoogeboom2022equivariantdiffusionmoleculegeneration} and attribute this to exhaustive learning rate tuning.}. In addition, we observe that energy loss is much more data efficient than baselines, allowing for the training of capable molecular generative models, producing over $75\%$ stable molecules using only $50\%$ of the training set (50K samples). 
\cref{tab:geom_drugs} contains results on the GEOM-Drugs data. \cref{tab:qm9} shows results on the QM9 data with GDM-aug model and and its equivariant variant EDM, with comprehensive results in \cref{apd:molecule}. Importantly, \autoref{tab:qm9} shows that energy loss with a non-equivariant architecture results in more improvement than using an equivariant architecture, at negligible computational cost.

\label{subsec:molecule}

\begin{table}[t]
\centering
\scriptsize
\caption{3D and alignment metrics for JODO variants.}
\label{tab:mol_metrics}
\begin{tabular}{@{}lccccccccc@{}}
  \toprule
  & \multicolumn{5}{c}{\textbf{Metric-3D}} & \multicolumn{3}{c}{\textbf{Metric-Align}} \\
  \cmidrule(lr){2-6} \cmidrule(lr){7-9}
  Model & At. stab. (\%) & Mol. stab. (\%) & Val. (\%) & Compl. (\%) & FCD ↓ &
  Bond ↓ & Angle ↓ & Dihedral ↓ \\
  \midrule
  JODO (paper) & 99.2 & 93.4 & — & — & 0.885 &
  0.1475 & 0.0121 & 6.29e-4 \\
  JODO (ours) & 99.2 & 92.8 & 95.6 & 95.5 & \textbf{0.854} &
  0.1218 & 0.0110 & 5.91e-4 \\
  JODO + Energy (Inv.) & 99.4 & 94.3 & 97.1 & 97.0 & 0.892 &
  0.1125 & \textbf{0.0046} & \textbf{4.95e-4} \\
  JODO + Energy (Exp.) & \textbf{99.6} & \textbf{96.6} & \textbf{98.4} & \textbf{98.4} & 1.495 &
  \textbf{0.0928} & 0.0142 & 4.97e-3 \\
  \bottomrule
\end{tabular}
\end{table}

The results using the JODO model are reported in \cref{tab:mol_metrics}. We observe that using energy loss with JODO is able to improve all align metrics, and nearly all 3D metrics, with comparable FCD, compared to the default Kabsch-aligned loss. This suggests that energy loss can push the state of the art and offers complementary benefits to equivariant architectures. 

\begin{wraptable}{r}{6.5cm}
\vspace{-3ex}
\scriptsize
\caption{Ablation for coefficients $k_{ij}\pr{\v{y}}$.}\label{tab:coeff}
\begin{tabular}{@{}lccc@{}}
  \toprule
  Coeff. & Mol. stab. (\%) & Atom stab. (\%) & Valid. (\%) \\
  \midrule
  Exp. Dist.      & \textbf{89.8} $\pm$ 2.8 & \textbf{99.3} $\pm$ 0.3 & \textbf{97.7} $\pm$ 1.4 \\
  Inv. Sq. Dist.  & 84.6 $\pm$ 1.8 & 98.9 $\pm$ 0.2 & 96.6 $\pm$ 1.5 \\
  Inv. Dist.      & 84.5 $\pm$ 2.1 & 98.7 $\pm$ 0.2 & 95.0 $\pm$ 1.5 \\
  Constant        & 83.6 $\pm$ 1.5 & 98.7 $\pm$ 0.1 & 93.6 $\pm$ 0.7 \\
  \bottomrule
\end{tabular}
\end{wraptable}
\paragraph{Ablation.}
We conduct an ablation over the form of the spring coefficients $k_{ij}\pr{\v{y}}$ in the energy loss (\cref{tab:coeff}). We consider the following functional forms: constant, inverse distance, inverse square distance, and exponential decay. A thorough sweep over learning rates was conducted. With EDM/GDM, we find exponential decay to give the best empirical results.
However, inverse distance coefficients work well for JODO and \cref{apd:sparse_timing} shows a less stark decay works better for the sparse loss function on large molecules. This suggests it is necessary to ablate these coefficients on new tasks.

\vspace{-0.25ex}
\subsection{Spin ground state prediction}

\textbf{Experimental Setup}
We consider the task of predicting the ground states of the spin Hamiltonian and compare the effectiveness of different loss functions. We construct a dataset of 10,000 training and test spin-glass Hamiltonians, each with couplings uniformly sampled from $\br{-1,1}$. We consider grids of size $16 \times 16$, which offer a challenging problem. The target ground-states are obtained by solving the associated integer linear program \citep{billionnet2007using}.
A convolutional neural network (CNN) is trained to take as input the coefficients $J_{ij}$ and predict the ground state configuration $\hat{\v{y}}_i$. More details on the architecture and training setup are provided in \cref{apd:spin}. We compare training with the energy loss function of \cref{{eq:spinloss}} to the cross-entropy loss function and the margin loss function, another commonly used loss function for classification. The evaluation metric we consider is the energy of the predicted configuration. We also compare with direct minimization of the true energy as a baseline, despite it not being a classification objective.
\begin{wraptable}{r}{10cm}
\centering
{\footnotesize
\caption{Results on ground-state prediction.}\label{tab:spins}
\begin{tabular}{llll|l}
  \toprule
  Loss & Cross-entropy & Margin loss & Local energy & True energy \\
  \midrule
  Test energy & 58.8 $\pm$ 0.8 & 49.87 $\pm$ 1.5 & 45.6 $\pm$ 1.6 & 14.6 $\pm$ 0.3 \\
  \bottomrule
\end{tabular}}
\end{wraptable}

\textbf{Results}
The results in \cref{tab:spins} show that using the local energy leads to lower configuration energies than the cross-entropy loss function and the margin loss function. As expected, minimizing the true energy still leads to lower overall energy, despite not using the data. The local-field loss also requires fewer training epochs to converge than the cross-entropy loss.
The results support the hypothesis that directly embedding physical insights through the local-field formulation effectively guides the learning process toward physically meaningful predictions.

\vspace{-1ex}
\section{Conclusion}
\label{sec:conclusion}
We demonstrated a new approach to designing loss functions for machine learning tasks in physical systems based on the system's energy.
When applied to both continuous and discrete settings, we found that replacing a simple MSE or cross-entropy loss with our energy loss functions leads to improved predictions across experiments. We further demonstrate the suitability of this loss for diffusion models and analyze its symmetry-invariance properties and scalability.

\paragraph{Limitations and future work} Some limitations remain, which also point to directions for future work. First, when energy loss functions are used for diffusion models, the correct score is recovered at low noise levels; exact recovery at higher noise levels would require an explicit correction, which we leave for later study. Second, while we offer a more principled approach to designing loss functions, some choices are still ad hoc. Looking ahead, richer surrogate energies that capture torsional angles could be investigated. The approach could also potentially be fruitfully extended to a broader class of systems, including crystalline materials and proteins.

\section*{Acknowledgments}
We are thankful to Mohsin Hasan, Simon Blackburn, Bruno Rousseau and Simon Verret for helpful discussions. 
This research was supported by the CIFAR AI Chairs program, Intel AI Labs and NSERC Discovery. S.-O. K.'s research is also supported by IVADO and FRQNT, and D.L. research is additionally supported by FRQNT. Mila and Compute Canada provided computational resources. 

\bibliography{biblio}

\begin{thebibliography}{76}
\providecommand{\natexlab}[1]{#1}
\providecommand{\url}[1]{\texttt{#1}}
\expandafter\ifx\csname urlstyle\endcsname\relax
  \providecommand{\doi}[1]{doi: #1}\else
  \providecommand{\doi}{doi: \begingroup \urlstyle{rm}\Url}\fi

\bibitem[Abramson et~al.(2024)Abramson, Adler, Dunger, Evans, Green, Pritzel,
  Ronneberger, Willmore, Ballard, Bambrick, et~al.]{abramson2024accurate}
Josh Abramson, Jonas Adler, Jack Dunger, Richard Evans, Tim Green, Alexander
  Pritzel, Olaf Ronneberger, Lindsay Willmore, Andrew~J Ballard, Joshua
  Bambrick, et~al.
\newblock Accurate structure prediction of biomolecular interactions with
  alphafold 3.
\newblock \emph{Nature}, pages 1--3, 2024.

\bibitem[Albergo et~al.(2023)Albergo, Boffi, and
  Vanden-Eijnden]{albergo2023stochastic}
Michael~S Albergo, Nicholas~M Boffi, and Eric Vanden-Eijnden.
\newblock Stochastic interpolants: A unifying framework for flows and
  diffusions.
\newblock \emph{arXiv preprint arXiv:2303.08797}, 2023.

\bibitem[Asimow and Roth(1978)]{asimow1978rigidity}
Leonard Asimow and Ben Roth.
\newblock The rigidity of graphs.
\newblock \emph{Transactions of the American Mathematical Society},
  245:\penalty0 279--289, 1978.

\bibitem[Axelrod and
  Gomez-Bombarelli(2022)]{axelrod2022geomenergyannotatedmolecularconformations}
Simon Axelrod and Rafael Gomez-Bombarelli.
\newblock Geom: Energy-annotated molecular conformations for property
  prediction and molecular generation, 2022.
\newblock URL \url{https://arxiv.org/abs/2006.05531}.

\bibitem[Bastek et~al.(2024)Bastek, Sun, and Kochmann]{bastek2024physics}
Jan-Hendrik Bastek, WaiChing Sun, and Dennis~M Kochmann.
\newblock Physics-informed diffusion models.
\newblock \emph{arXiv preprint arXiv:2403.14404}, 2024.

\bibitem[Billionnet and Elloumi(2007)]{billionnet2007using}
Alain Billionnet and Sourour Elloumi.
\newblock Using a mixed integer quadratic programming solver for the
  unconstrained quadratic 0-1 problem.
\newblock \emph{Mathematical programming}, 109:\penalty0 55--68, 2007.

\bibitem[Carrasquilla and Melko(2017)]{carrasquilla2017}
Juan Carrasquilla and Roger~G. Melko.
\newblock Machine learning phases of matter.
\newblock \emph{Nature Physics}, 13\penalty0 (5):\penalty0 431--434, 2017.
\newblock \doi{10.1038/nphys4035}.
\newblock URL \url{https://doi.org/10.1038/nphys4035}.

\bibitem[Chandrasekhar(1943)]{chandrasekhar1943stochastic}
Subrahmanyan Chandrasekhar.
\newblock Stochastic problems in physics and astronomy.
\newblock \emph{Reviews of modern physics}, 15\penalty0 (1):\penalty0 1, 1943.

\bibitem[Cognolato et~al.(2025)Cognolato, Rigoni, Ballarini, Serafini, Moro,
  Sperduti, et~al.]{cognolato2025d4}
Samuel Cognolato, Davide Rigoni, Marco Ballarini, Luciano Serafini, Stefano
  Moro, Alessandro Sperduti, et~al.
\newblock D4: Distance diffusion for a truly equivariant molecular design.
\newblock In \emph{ESANN 2025-Proceedings, 33rd European Symposium on
  Artificial Neural Networks, Computational Intelligence and Machine Learning},
  pages 265--270, 2025.

\bibitem[Cranmer et~al.(2020)Cranmer, Greydanus, Hoyer, Battaglia, Spergel, and
  Ho]{cranmer2020lagrangian}
Miles Cranmer, Sam Greydanus, Stephan Hoyer, Peter Battaglia, David Spergel,
  and Shirley Ho.
\newblock Lagrangian neural networks.
\newblock \emph{arXiv preprint arXiv:2003.04630}, 2020.

\bibitem[Dewar(2025)]{dewar2024rigidity}
Sean Dewar.
\newblock graph-rigidity-checker library.
\newblock \url{https://github.com/dewar28/graph-rigidity-checker/}, 2025.

\bibitem[Du and Mordatch(2019)]{du2019implicit}
Yilun Du and Igor Mordatch.
\newblock Implicit generation and modeling with energy based models.
\newblock \emph{Advances in neural information processing systems}, 32, 2019.

\bibitem[Einstein(1905)]{einstein1905motion}
Albert Einstein.
\newblock On the motion of small particles suspended in liquids at rest
  required by the molecular-kinetic theory of heat.
\newblock \emph{Annalen der physik}, 17\penalty0 (549-560):\penalty0 208, 1905.

\bibitem[Frauenfelder et~al.(1991)Frauenfelder, Sligar, and
  Wolynes]{frauenfelder1991energy}
Hans Frauenfelder, Stephen~G Sligar, and Peter~G Wolynes.
\newblock The energy landscapes and motions of proteins.
\newblock \emph{Science}, 254\penalty0 (5038):\penalty0 1598--1603, 1991.

\bibitem[Fu and Anderson(1986)]{fu1986}
Yaotian Fu and Philip~W Anderson.
\newblock Application of statistical mechanics to np-complete problems in
  combinatorial optimisation.
\newblock \emph{Journal of Physics A: Mathematical and General}, 19\penalty0
  (9):\penalty0 1605, 1986.

\bibitem[Gardiner(1985)]{gardiner1985handbook}
Crispin~W Gardiner.
\newblock Handbook of stochastic methods for physics, chemistry and the natural
  sciences.
\newblock \emph{Springer series in synergetics}, 1985.

\bibitem[G{\'o}mez-Bombarelli et~al.(2018)G{\'o}mez-Bombarelli, Wei, Duvenaud,
  Hern{\'a}ndez-Lobato, S{\'a}nchez-Lengeling, Sheberla, Aguilera-Iparraguirre,
  Hirzel, Adams, and Aspuru-Guzik]{gomez2018automatic}
Rafael G{\'o}mez-Bombarelli, Jennifer~N Wei, David Duvenaud, Jos{\'e}~Miguel
  Hern{\'a}ndez-Lobato, Benjam{\'\i}n S{\'a}nchez-Lengeling, Dennis Sheberla,
  Jorge Aguilera-Iparraguirre, Timothy~D Hirzel, Ryan~P Adams, and Al{\'a}n
  Aspuru-Guzik.
\newblock Automatic chemical design using a data-driven continuous
  representation of molecules.
\newblock \emph{ACS central science}, 4\penalty0 (2):\penalty0 268--276, 2018.

\bibitem[Grathwohl et~al.(2019)Grathwohl, Wang, Jacobsen, Duvenaud, Norouzi,
  and Swersky]{grathwohl2019your}
Will Grathwohl, Kuan-Chieh Wang, J{\"o}rn-Henrik Jacobsen, David Duvenaud,
  Mohammad Norouzi, and Kevin Swersky.
\newblock Your classifier is secretly an energy based model and you should
  treat it like one.
\newblock \emph{arXiv preprint arXiv:1912.03263}, 2019.

\bibitem[Greydanus et~al.(2019)Greydanus, Dzamba, and
  Yosinski]{greydanus2019hamiltonian}
Samuel Greydanus, Misko Dzamba, and Jason Yosinski.
\newblock Hamiltonian neural networks.
\newblock \emph{Advances in neural information processing systems}, 32, 2019.

\bibitem[Hassan et~al.(2024)Hassan, Shenoy, Lee, St{\"a}rk, Thaler, and
  Beaini]{hassan2024flow}
Majdi Hassan, Nikhil Shenoy, Jungyoon Lee, Hannes St{\"a}rk, Stephan Thaler,
  and Dominique Beaini.
\newblock Et-flow: Equivariant flow-matching for molecular conformer
  generation.
\newblock \emph{Advances in Neural Information Processing Systems},
  37:\penalty0 128798--128824, 2024.

\bibitem[Ho et~al.(2020)Ho, Jain, and Abbeel]{ho2020}
Jonathan Ho, Ajay Jain, and Pieter Abbeel.
\newblock Denoising diffusion probabilistic models.
\newblock In H.~Larochelle, M.~Ranzato, R.~Hadsell, M.F. Balcan, and H.~Lin,
  editors, \emph{Advances in Neural Information Processing Systems}, volume~33,
  pages 6840--6851. Curran Associates, Inc., 2020.
\newblock URL
  \url{https://proceedings.neurips.cc/paper_files/paper/2020/file/4c5bcfec8584af0d967f1ab10179ca4b-Paper.pdf}.

\bibitem[Hoogeboom et~al.(2022)Hoogeboom, Satorras, Vignac, and
  Welling]{hoogeboom2022equivariantdiffusionmoleculegeneration}
Emiel Hoogeboom, Victor~Garcia Satorras, Clément Vignac, and Max Welling.
\newblock Equivariant diffusion for molecule generation in 3d, 2022.
\newblock URL \url{https://arxiv.org/abs/2203.17003}.

\bibitem[Huang et~al.(2023)Huang, Sun, Du, and Lv]{huang2023learningjoint2d}
Han Huang, Leilei Sun, Bowen Du, and Weifeng Lv.
\newblock Learning joint 2d \& 3d diffusion models for complete molecule
  generation, 2023.
\newblock URL \url{https://arxiv.org/abs/2305.12347}.

\bibitem[Jaynes(1957)]{jaynes1957information}
Edwin~T Jaynes.
\newblock Information theory and statistical mechanics.
\newblock \emph{Physical review}, 106\penalty0 (4):\penalty0 620, 1957.

\bibitem[Jiao et~al.(2023)Jiao, Huang, Lin, Han, Chen, Lu, and
  Liu]{jiao2023crystal}
Rui Jiao, Wenbing Huang, Peijia Lin, Jiaqi Han, Pin Chen, Yutong Lu, and Yang
  Liu.
\newblock Crystal structure prediction by joint equivariant diffusion.
\newblock \emph{Advances in Neural Information Processing Systems},
  36:\penalty0 17464--17497, 2023.

\bibitem[Jing et~al.(2024)Jing, Berger, and Jaakkola]{jing2024alphafold}
Bowen Jing, Bonnie Berger, and Tommi Jaakkola.
\newblock Alphafold meets flow matching for generating protein ensembles.
\newblock \emph{arXiv preprint arXiv:2402.04845}, 2024.

\bibitem[Jumper et~al.(2021)Jumper, Evans, Pritzel, Green, Figurnov,
  Ronneberger, Tunyasuvunakool, Bates, {\v{Z}}{\'\i}dek, Potapenko,
  et~al.]{jumper2021highly}
John Jumper, Richard Evans, Alexander Pritzel, Tim Green, Michael Figurnov,
  Olaf Ronneberger, Kathryn Tunyasuvunakool, Russ Bates, Augustin
  {\v{Z}}{\'\i}dek, Anna Potapenko, et~al.
\newblock Highly accurate protein structure prediction with alphafold.
\newblock \emph{nature}, 596\penalty0 (7873):\penalty0 583--589, 2021.

\bibitem[Kaba and Ravanbakhsh(2023)]{kaba2023symmetry}
S{\'e}kou-Oumar Kaba and Siamak Ravanbakhsh.
\newblock Symmetry breaking and equivariant neural networks.
\newblock \emph{arXiv preprint arXiv:2312.09016}, 2023.

\bibitem[Kaba et~al.(2023)Kaba, Mondal, Zhang, Bengio, and
  Ravanbakhsh]{kaba2023equivariance}
S{\'e}kou-Oumar Kaba, Arnab~Kumar Mondal, Yan Zhang, Yoshua Bengio, and Siamak
  Ravanbakhsh.
\newblock Equivariance with learned canonicalization functions.
\newblock In \emph{International Conference on Machine Learning}, pages
  15546--15566. PMLR, 2023.

\bibitem[Kabsch(1976)]{kabsch1976solution}
Wolfgang Kabsch.
\newblock A solution for the best rotation to relate two sets of vectors.
\newblock \emph{Foundations of Crystallography}, 32\penalty0 (5):\penalty0
  922--923, 1976.

\bibitem[Kingma and Gao(2024)]{kingma2024understanding}
Diederik Kingma and Ruiqi Gao.
\newblock Understanding diffusion objectives as the elbo with simple data
  augmentation.
\newblock \emph{Advances in Neural Information Processing Systems}, 36, 2024.

\bibitem[Klein and Noé(2025)]{klein2025transferableboltzmanngenerators}
Leon Klein and Frank Noé.
\newblock Transferable boltzmann generators, 2025.
\newblock URL \url{https://arxiv.org/abs/2406.14426}.

\bibitem[Klein et~al.(2023)Klein, Krämer, and
  Noé]{klein2023equivariantflowmatching}
Leon Klein, Andreas Krämer, and Frank Noé.
\newblock Equivariant flow matching, 2023.
\newblock URL \url{https://arxiv.org/abs/2306.15030}.

\bibitem[K{\"o}hler et~al.(2020)K{\"o}hler, Klein, and
  No{\'e}]{kohler2020equivariant}
Jonas K{\"o}hler, Leon Klein, and Frank No{\'e}.
\newblock Equivariant flows: exact likelihood generative learning for symmetric
  densities.
\newblock In \emph{International conference on machine learning}, pages
  5361--5370. PMLR, 2020.

\bibitem[Kramers(1940)]{kramers1940brownian}
Hendrik~Anthony Kramers.
\newblock Brownian motion in a field of force and the diffusion model of
  chemical reactions.
\newblock \emph{physica}, 7\penalty0 (4):\penalty0 284--304, 1940.

\bibitem[Krivelevich et~al.(2023)Krivelevich, Lew, and
  Michaeli]{krivelevich2023rigid}
Michael Krivelevich, Alan Lew, and Peleg Michaeli.
\newblock Rigid partitions: from high connectivity to random graphs.
\newblock \emph{arXiv preprint arXiv:2311.14451}, 2023.

\bibitem[Kurtzberg(1962)]{kurtzberg1962approximation}
Jerome~M Kurtzberg.
\newblock On approximation methods for the assignment problem.
\newblock \emph{Journal of the ACM (JACM)}, 9\penalty0 (4):\penalty0 419--439,
  1962.

\bibitem[Laman(1970)]{laman1970graphs}
Gerard Laman.
\newblock On graphs and rigidity of plane skeletal structures.
\newblock \emph{Journal of Engineering mathematics}, 4\penalty0 (4):\penalty0
  331--340, 1970.

\bibitem[Langevin(1908)]{langevin1908theory}
Paul Langevin.
\newblock On the theory of brownian motion.
\newblock \emph{CR Acad. Sci. Paris}, 146\penalty0 (530-533):\penalty0 530,
  1908.

\bibitem[Lawrence et~al.(2025)Lawrence, Portilheiro, Zhang, and
  Kaba]{lawrence2025improving}
Hannah Lawrence, Vasco Portilheiro, Yan Zhang, and S{\'e}kou-Oumar Kaba.
\newblock Improving equivariant networks with probabilistic symmetry breaking.
\newblock In \emph{The Thirteenth International Conference on Learning
  Representations}, 2025.
\newblock URL \url{https://openreview.net/forum?id=ZE6lrLvATd}.

\bibitem[LeCun et~al.(2006)LeCun, Chopra, Hadsell, Ranzato, Huang,
  et~al.]{lecun2006tutorial}
Yann LeCun, Sumit Chopra, Raia Hadsell, M~Ranzato, Fujie Huang, et~al.
\newblock A tutorial on energy-based learning.
\newblock \emph{Predicting structured data}, 1\penalty0 (0), 2006.

\bibitem[Lennard-Jones(1931)]{lennard-jones}
J~E Lennard-Jones.
\newblock Cohesion.
\newblock \emph{Proceedings of the Physical Society}, 43\penalty0 (5):\penalty0
  461, 1931.
\newblock \doi{10.1088/0959-5309/43/5/301}.
\newblock URL \url{https://dx.doi.org/10.1088/0959-5309/43/5/301}.

\bibitem[Lew et~al.(2023)Lew, Nevo, Peled, and Raz]{blms.12740}
Alan Lew, Eran Nevo, Yuval Peled, and Orit~E. Raz.
\newblock Sharp threshold for rigidity of random graphs.
\newblock \emph{Bulletin of the London Mathematical Society}, 55\penalty0
  (1):\penalty0 490--501, 2023.
\newblock \doi{https://doi.org/10.1112/blms.12740}.

\bibitem[Lipman et~al.(2023)Lipman, Chen, Ben-Hamu, Nickel, and
  Le]{lipman2023flow}
Yaron Lipman, Ricky T.~Q. Chen, Heli Ben-Hamu, Maximilian Nickel, and Matthew
  Le.
\newblock Flow matching for generative modeling.
\newblock In \emph{The Eleventh International Conference on Learning
  Representations}, 2023.
\newblock URL \url{https://openreview.net/forum?id=PqvMRDCJT9t}.

\bibitem[M{\'e}zard et~al.(1987)M{\'e}zard, Parisi, and
  Virasoro]{mezard1987spin}
Marc M{\'e}zard, Giorgio Parisi, and Miguel~Angel Virasoro.
\newblock \emph{Spin glass theory and beyond: An Introduction to the Replica
  Method and Its Applications}, volume~9.
\newblock World Scientific Publishing Company, 1987.

\bibitem[Morse(1929)]{morse1929diatomic}
Philip~M Morse.
\newblock Diatomic molecules according to the wave mechanics. ii. vibrational
  levels.
\newblock \emph{Physical review}, 34\penalty0 (1):\penalty0 57, 1929.

\bibitem[Nesterov et~al.(2020)Nesterov, Wieser, and Roth]{nesterov20203dmolnet}
Vitali Nesterov, Mario Wieser, and Volker Roth.
\newblock 3dmolnet: a generative network for molecular structures.
\newblock \emph{arXiv preprint arXiv:2010.06477}, 2020.

\bibitem[No{\'e} et~al.(2019)No{\'e}, Olsson, K{\"o}hler, and
  Wu]{noe2019boltzmann}
Frank No{\'e}, Simon Olsson, Jonas K{\"o}hler, and Hao Wu.
\newblock Boltzmann generators: Sampling equilibrium states of many-body
  systems with deep learning.
\newblock \emph{Science}, 365\penalty0 (6457):\penalty0 eaaw1147, 2019.

\bibitem[No{\'e} et~al.(2020)No{\'e}, De~Fabritiis, and
  Clementi]{noe2020machine}
Frank No{\'e}, Gianni De~Fabritiis, and Cecilia Clementi.
\newblock Machine learning for protein folding and dynamics.
\newblock \emph{Current opinion in structural biology}, 60:\penalty0 77--84,
  2020.

\bibitem[Pahng and Brenner(2020)]{pahng2020predicting}
Seong~Ho Pahng and Michael~P Brenner.
\newblock Predicting ground state configuration of energy landscape ensemble
  using graph neural network.
\newblock \emph{arXiv preprint arXiv:2008.08227}, 2020.

\bibitem[Pathria(2017)]{pathria2017statistical}
Raj~Kumar Pathria.
\newblock \emph{Statistical Mechanics: International Series of Monographs in
  Natural Philosophy}, volume~45.
\newblock Elsevier, 2017.

\bibitem[Peled(2024)]{rigidity_talk}
Yuval Peled.
\newblock Sharp threshold for rigidity of random graphs.
\newblock Presented at IPAM at UCLA, 2024.

\bibitem[Peng et~al.(2023)Peng, Guan, Liu, and Ma]{peng2023moldiff}
Xingang Peng, Jiaqi Guan, Qiang Liu, and Jianzhu Ma.
\newblock Moldiff: Addressing the atom-bond inconsistency problem in 3d
  molecule diffusion generation.
\newblock \emph{arXiv preprint arXiv:2305.07508}, 2023.

\bibitem[Preuer et~al.(2018)Preuer, Renz, Unterthiner, Hochreiter, and
  Klambauer]{fcd2018}
Kristina Preuer, Philipp Renz, Thomas Unterthiner, Sepp Hochreiter, and Günter
  Klambauer.
\newblock Fréchet chemnet distance: A metric for generative models for
  molecules in drug discovery.
\newblock \emph{Journal of Chemical Information and Modeling}, 58\penalty0
  (9):\penalty0 1736--1741, 2018.
\newblock \doi{10.1021/acs.jcim.8b00234}.

\bibitem[Raissi et~al.(2019)Raissi, Perdikaris, and
  Karniadakis]{raissi2019physics}
Maziar Raissi, Paris Perdikaris, and George~E Karniadakis.
\newblock Physics-informed neural networks: A deep learning framework for
  solving forward and inverse problems involving nonlinear partial differential
  equations.
\newblock \emph{Journal of Computational physics}, 378:\penalty0 686--707,
  2019.

\bibitem[Ramakrishnan et~al.(2014)Ramakrishnan, Dral, Rupp, and von
  Lilienfeld]{Ramakrishnan2014}
Raghunathan Ramakrishnan, Pavlo~O. Dral, Matthias Rupp, and O.~Anatole von
  Lilienfeld.
\newblock Quantum chemistry structures and properties of 134 kilo molecules.
\newblock \emph{Scientific Data}, 1\penalty0 (1):\penalty0 140022, 2014.
\newblock ISSN 2052-4463.
\newblock \doi{10.1038/sdata.2014.22}.
\newblock URL \url{https://doi.org/10.1038/sdata.2014.22}.

\bibitem[Ryan et~al.(2018)Ryan, Lengyel, and Shatruk]{ryan2018crystal}
Kevin Ryan, Jeff Lengyel, and Michael Shatruk.
\newblock Crystal structure prediction via deep learning.
\newblock \emph{Journal of the American Chemical Society}, 140\penalty0
  (32):\penalty0 10158--10168, 2018.

\bibitem[Sanchez-Lengeling and Aspuru-Guzik(2018)]{sanchez2018inverse}
Benjamin Sanchez-Lengeling and Al{\'a}n Aspuru-Guzik.
\newblock Inverse molecular design using machine learning: Generative models
  for matter engineering.
\newblock \emph{Science}, 361\penalty0 (6400):\penalty0 360--365, 2018.

\bibitem[Sareen et~al.(2025)Sareen, Levy, Mondal, Kaba, Akhound-Sadegh, and
  Ravanbakhsh]{sareen2025symmetryawaregenerativemodelinglearned}
Kusha Sareen, Daniel Levy, Arnab~Kumar Mondal, Sékou-Oumar Kaba, Tara
  Akhound-Sadegh, and Siamak Ravanbakhsh.
\newblock Symmetry-aware generative modeling through learned canonicalization,
  2025.
\newblock URL \url{https://arxiv.org/abs/2501.07773}.

\bibitem[Satorras et~al.(2022)Satorras, Hoogeboom, Fuchs, Posner, and
  Welling]{satorras2022enequivariantnormalizingflows}
Victor~Garcia Satorras, Emiel Hoogeboom, Fabian~B. Fuchs, Ingmar Posner, and
  Max Welling.
\newblock E(n) equivariant normalizing flows, 2022.
\newblock URL \url{https://arxiv.org/abs/2105.09016}.

\bibitem[Schuch and Verstraete(2009)]{schuch2009computational}
Norbert Schuch and Frank Verstraete.
\newblock Computational complexity of interacting electrons and fundamental
  limitations of density functional theory.
\newblock \emph{Nature physics}, 5\penalty0 (10):\penalty0 732--735, 2009.

\bibitem[Simm and Hern{\'a}ndez-Lobato(2019)]{simm2019generative}
Gregor~NC Simm and Jos{\'e}~Miguel Hern{\'a}ndez-Lobato.
\newblock A generative model for molecular distance geometry.
\newblock \emph{arXiv preprint arXiv:1909.11459}, 2019.

\bibitem[Smidt et~al.(2021)Smidt, Geiger, and Miller]{smidt2021finding}
Tess~E Smidt, Mario Geiger, and Benjamin~Kurt Miller.
\newblock Finding symmetry breaking order parameters with euclidean neural
  networks.
\newblock \emph{Physical Review Research}, 3\penalty0 (1):\penalty0 L012002,
  2021.

\bibitem[Sohl-Dickstein et~al.(2015)Sohl-Dickstein, Weiss, Maheswaranathan, and
  Ganguli]{dickstein2015}
Jascha Sohl-Dickstein, Eric Weiss, Niru Maheswaranathan, and Surya Ganguli.
\newblock Deep unsupervised learning using nonequilibrium thermodynamics.
\newblock In Francis Bach and David Blei, editors, \emph{Proceedings of the
  32nd International Conference on Machine Learning}, volume~37 of
  \emph{Proceedings of Machine Learning Research}, pages 2256--2265, Lille,
  France, 07--09 Jul 2015. PMLR.
\newblock URL \url{https://proceedings.mlr.press/v37/sohl-dickstein15.html}.

\bibitem[Song et~al.(2021)Song, Sohl-Dickstein, Kingma, Kumar, Ermon, and
  Poole]{song2021scorebased}
Yang Song, Jascha Sohl-Dickstein, Diederik~P Kingma, Abhishek Kumar, Stefano
  Ermon, and Ben Poole.
\newblock Score-based generative modeling through stochastic differential
  equations.
\newblock In \emph{International Conference on Learning Representations}, 2021.
\newblock URL \url{https://openreview.net/forum?id=PxTIG12RRHS}.

\bibitem[Thorpe and Duxbury(1999)]{thorpe1999rigidity}
M.F. Thorpe and P.M. Duxbury.
\newblock \emph{Rigidity Theory and Applications}.
\newblock Fundamental Materials Research. Springer US, 1999.
\newblock ISBN 9780306461156.
\newblock URL \url{https://books.google.ca/books?id=3XgykKbZymkC}.

\bibitem[Villar et~al.(2021)Villar, Hogg, Storey-Fisher, Yao, and
  Blum-Smith]{villar2021scalars}
Soledad Villar, David~W Hogg, Kate Storey-Fisher, Weichi Yao, and Ben
  Blum-Smith.
\newblock Scalars are universal: Equivariant machine learning, structured like
  classical physics.
\newblock In A.~Beygelzimer, Y.~Dauphin, P.~Liang, and J.~Wortman Vaughan,
  editors, \emph{Advances in Neural Information Processing Systems}, 2021.

\bibitem[Vincent(2011)]{vincent2011connection}
Pascal Vincent.
\newblock A connection between score matching and denoising autoencoders.
\newblock \emph{Neural computation}, 23\penalty0 (7):\penalty0 1661--1674,
  2011.

\bibitem[Wales et~al.(2000)Wales, Doye, Miller, Mortenson, and
  Walsh]{wales2000energy}
David~J Wales, Jonathan~PK Doye, Mark~A Miller, Paul~N Mortenson, and Tiffany~R
  Walsh.
\newblock Energy landscapes: from clusters to biomolecules.
\newblock \emph{Advances in Chemical Physics}, 115:\penalty0 1--111, 2000.

\bibitem[Wang et~al.(2025)Wang, Huang, Baker, Sun, Tang, and Wang]{wang2025a}
Shih-Hsin Wang, Yuhao Huang, Justin~M. Baker, Yuan-En Sun, Qi~Tang, and Bao
  Wang.
\newblock A theoretically-principled sparse, connected, and rigid graph
  representation of molecules.
\newblock In \emph{The Thirteenth International Conference on Learning
  Representations}, 2025.
\newblock URL \url{https://openreview.net/forum?id=OIvg3MqWX2}.

\bibitem[Xie and Smidt(2024)]{xie2024equivariant}
YuQing Xie and Tess Smidt.
\newblock Equivariant symmetry breaking sets.
\newblock \emph{Transactions on Machine Learning Research}, 2024.
\newblock ISSN 2835-8856.
\newblock URL \url{https://openreview.net/forum?id=tHKH4DNSR5}.

\bibitem[Xu et~al.(2021)Xu, Luo, Bengio, Peng, and Tang]{xu2021learning}
Minkai Xu, Shitong Luo, Yoshua Bengio, Jian Peng, and Jian Tang.
\newblock Learning neural generative dynamics for molecular conformation
  generation.
\newblock \emph{arXiv preprint arXiv:2102.10240}, 2021.

\bibitem[Yang and G{\'o}mez-Bombarelli(2023)]{yang2023chemically}
Soojung Yang and Rafael G{\'o}mez-Bombarelli.
\newblock Chemically transferable generative backmapping of coarse-grained
  proteins.
\newblock \emph{arXiv preprint arXiv:2303.01569}, 2023.

\bibitem[Zeni et~al.(2025)Zeni, Pinsler, Z{\"u}gner, Fowler, Horton, Fu, Wang,
  Shysheya, Crabb{\'e}, Ueda, et~al.]{zeni2025generative}
Claudio Zeni, Robert Pinsler, Daniel Z{\"u}gner, Andrew Fowler, Matthew Horton,
  Xiang Fu, Zilong Wang, Aliaksandra Shysheya, Jonathan Crabb{\'e}, Shoko Ueda,
  et~al.
\newblock A generative model for inorganic materials design.
\newblock \emph{Nature}, pages 1--3, 2025.

\bibitem[Zhang et~al.(2024)Zhang, Ashouritaklimi, Teh, and
  Cornish]{zhang2024symdiff}
Leo Zhang, Kianoosh Ashouritaklimi, Yee~Whye Teh, and Rob Cornish.
\newblock Symdiff: Equivariant diffusion via stochastic symmetrisation.
\newblock \emph{arXiv preprint arXiv:2410.06262}, 2024.

\bibitem[Zhang et~al.(2023)Zhang, Wang, Helwig, Luo, Fu, Xie, Liu, Lin, Xu,
  Yan, et~al.]{zhang2023artificial}
Xuan Zhang, Limei Wang, Jacob Helwig, Youzhi Luo, Cong Fu, Yaochen Xie, Meng
  Liu, Yuchao Lin, Zhao Xu, Keqiang Yan, et~al.
\newblock Artificial intelligence for science in quantum, atomistic, and
  continuum systems.
\newblock \emph{arXiv preprint arXiv:2307.08423}, 2023.

\end{thebibliography}
\bibliographystyle{plainnat}


%


\newpage

\newpage
\appendix

\section{Additional Theory}

\subsection{Proofs}
\label{apd:proofs}

\subsubsection{Proof of \cref{{prop:symmetry}}}

\begin{proof}
For any $\pr{g_1, g_2}\in E(d) \times (\mathrm{Aut}\pr{k\pr{\v{y}}}\cap \mathrm{Aut}\pr{\Delta y})$ where $g_1 \in E(d)$ and $g_2 \in (\mathrm{Aut}\pr{k\pr{\v{y}}}\cap \mathrm{Aut}\pr{\Delta y})$, we have
\begin{align}
& E\pr{\pr{g_1, g_2}\cdot \hat{\v{y}}, \v{y}} = \sum_{i,j}^n \frac{1}{2} k_{ij}\pr{\v{y}} \left(\lVert\v{y}_i -\v{y}_j \rVert - \lVert\pr{g_1, g_2}\cdot\hat{\v{y}}_i -\pr{g_1, g_2}\cdot\hat{\v{y}}_j \rVert \right)^2
\end{align}
By linearity of the actions, we have
\begin{align}
E\pr{\pr{g_1, g_2}\cdot \hat{\v{y}}, \v{y}} =  \sum_{i,j}^n \frac{1}{2} k_{ij}\pr{\v{y}} \left(\lVert\v{y}_i -\v{y}_j \rVert - \lVert g_1\cdot\pr{ g_2\cdot\hat{\v{y}}_i - g_2\cdot\hat{\v{y}}_j} \rVert \right)^2
\end{align}
Since the Euclidean norm of a difference is $E(d)$-invariant we have
\begin{align}
E\pr{\pr{g_1, g_2}\cdot \hat{\v{y}}, \v{y}} =  \sum_{i,j}^n \frac{1}{2} k_{ij}\pr{\v{y}} \left(\lVert\v{y}_i -\v{y}_j \rVert - \lVert \pr{g_2\cdot\hat{ \v{y}}_i - g_2\cdot\hat{\v{y}}_j} \rVert \right)^2
\end{align}
We then have
\begin{align}
E\pr{\pr{g_1, g_2}\cdot \hat{\v{y}}, \v{y}} =  \sum_{i,j}^n \frac{1}{2} k_{ij}\pr{\v{y}} \left(\lVert\v{y}_i -\v{y}_j \rVert - \lVert \pr{\hat{\v{y}}_{g_2^{-1}\cdot i} - \hat{\v{y}}_{g_2^{-1}\cdot j}} \rVert \right)^2
\end{align}
Using the fact that $g_2 \in (\mathrm{Aut}\pr{k\pr{\v{y}}}\cap \mathrm{Aut}\pr{\Delta y})$,
\begin{align}
E\pr{\pr{g_1, g_2}\cdot \hat{\v{y}}, \v{y}} &=  \sum_{i,j}^n \frac{1}{2} k_{g_2^{-1}\cdot i, g_2^{-1}\cdot j}\pr{\v{y}} \left(\lVert\v{y}_{g_2^{-1}\cdot i} -\v{y}_{g_2^{-1}\cdot j} \rVert - \lVert \pr{\hat{\v{y}}_{g_2^{-1}\cdot i} - \hat{\v{y}}_{g_2^{-1}\cdot j}} \rVert \right)^2\\
E\pr{\pr{g_1, g_2}\cdot \hat{\v{y}}, \v{y}} &=\sum_{i,j}^n g_2\cdot \frac{1}{2} \br{k_{ij}\pr{\v{y}} \left(\lVert\v{y}_i -\v{y}_j \rVert - \lVert \pr{\hat{ \v{y}}_i - \hat{\v{y}}_j} \rVert \right)^2}
\end{align}
Since the sum is permutation invariant, we have
\begin{align}
E\pr{\pr{g_1, g_2}\cdot \hat{\v{y}}, \v{y}} &=\sum_{i,j}^n  \frac{1}{2} \br{k_{ij}\pr{\v{y}} \left(\lVert\v{y}_i -\v{y}_j \rVert - \lVert \pr{\hat{ \v{y}}_i - \hat{\v{y}}_j} \rVert \right)^2}\\
E\pr{\pr{g_1, g_2}\cdot \hat{\v{y}}, \v{y}} &=E\pr{\hat{\v{y}}, \v{y}} 
\end{align}

For the second argument, we similarly have
\begin{align}
E\pr{\hat{\v{y}}, \pr{g_1, g_2}\cdot \v{y}} &=E\pr{\hat{\v{y}}, \v{y}},
\end{align}
due to the $E(d)$-invariance of the Euclidean norm and the automorphism of $\v{y}$.

This completes the proof.

\end{proof}

\subsubsection{Proof of \cref{corol:min}}

\begin{proof}
First, 
\begin{align}
E({\v{y}}, \v{y}) = 0 = \min_{\hat{\v{y}}\in \mathbb{R}^{n\times d}} E(\hat{\v{y}}, \v{y})
\end{align}
Then, given
\begin{align}
& E\pr{{\v{y}}, \v{y}} = \sum_{i,j}^n \frac{1}{2} k_{ij}\pr{\v{y}} \left(\lVert\v{y}_i -\v{y}_j \rVert - \lVert\hat{\v{y}}_i -\hat{\v{y}}_j \rVert \right)^2
\end{align}
we see that for any $k_{ij}\pr{\v{y}}>0$ the loss is minimized when each term of the sum is zero which implies that $\Delta y = \Delta \hat{y}$ when the loss is minimized.

This implies
\begin{align}
\argmin_{\hat{\v{y}}\in \mathbb{R}^{n\times d}} E(\hat{\v{y}}, \v{y}) = \cbrace{g\cdot \v{y} \ \mid \ g\in E(d)}.
\end{align}
Since the action of the group $\mathrm{Aut}\pr{k\pr{\v{y}}}\cap \mathrm{Aut}\pr{\Delta y}$ on $\v{y}$ is by definition trivial, the desired result follows.

\end{proof}

\subsubsection{Proof of \cref{prop:score}}
\begin{proof}
We first establish some preliminaries. Since $SE(d)$ is not compact, we do not assume that the density $p\pr{\v{x}_t}$ is normalized. This is not an issue since the results depend on the score $\nabla_{\v{x}_t} \log p\pr{\v{x}_t}$, which is independent of the normalization.

We also parametrize the score in terms of a noise prediction to align with practice.
The diffusion objective for time $t$ is proportional to (multiplication by constants do not change the minimizer)
\begin{align}
\mathcal{J}_t\pr{\theta} = \mathbb{E}_{{\v{x}\sim p(\v{x}), \bm{\epsilon}\sim p\pr{\bm{\epsilon}}}} \br{ \norm{\frac{1}{\alpha_t}\v{x}_t- \frac{\sigma_t}{\alpha_t}\hat{\bm{\epsilon}}_\theta  - \v{x}}^2}
\end{align}
where $\hat{\bm{\epsilon}}_\theta = f_{\theta}\pr{\v{x}_t, t}$, $\v{x}_t = \alpha_t \v{x} + \sigma_t \bm{\epsilon}$ and $p\pr{\bm{\epsilon}} = \mathcal{N}\pr{0, \v{I}}$. The sample prediction is given by $\hat{\v{x}}_{\theta} = \frac{\pr{\v{x}_t - \sigma_t \hat{\bm{\epsilon}}_{\theta}\pr{\v{x}_t, t}}}{\alpha_t}$.

The expectation can be rewritten as
\begin{align}
&\mathcal{J}_t\pr{\theta} = \mathbb{E}_{{\v{x}_t\sim p(\v{x}_t), \v{x}\sim p\pr{\v{x}\mid \v{x}_t}}} \br{ \norm{\frac{1}{\alpha_t}\v{x}_t- \frac{\sigma_t}{\alpha_t}\hat{\bm{\epsilon}}_\theta  - \v{x}}^2}\\
& \mathcal{J}_t\pr{\theta} = \mathbb{E}_{{\v{x}_t\sim p(\v{x}_t)}} \br{\mathbb{E}_{{\v{x}\sim p\pr{\v{x}\mid \v{x}_t}}}\br{ \norm{\frac{1}{\alpha_t}\v{x}_t- \frac{\sigma_t}{\alpha_t}\hat{\bm{\epsilon}}_\theta  - \v{x}}^2}}\\
& \mathcal{J}_t\pr{\theta} = \mathbb{E}_{{\v{x}_t\sim p(\v{x}_t)}} \br{\mathcal{J}_t\pr{\v{x}_t,\theta}}
\end{align}

The objective is minimized if for all $\v{x}_t$, we have the following noise prediction
\begin{align}
\hat{\bm{\epsilon}}_{\text{MSE}}^* = \argmin_{\hat{\bm{\epsilon}}_\theta \in \mathbb{R}^{n\times d}}\mathcal{J}_t\pr{\v{x}_t,\theta} = \argmin_{\hat{\bm{\epsilon}}_\theta \in \mathbb{R}^{n\times d}}\mathbb{E}_{{\v{x}\sim p\pr{\v{x}\mid \v{x}_t}}}\br{ \norm{\frac{1}{\alpha_t}\v{x}_t- \frac{\sigma_t}{\alpha_t}\hat{\bm{\epsilon}}_\theta  - \v{x}}^2}
\end{align}
Since expectation and minimization commute for the MSE, we have
\begin{align}
&\hat{\bm{\epsilon}}_{\text{MSE}}^* = \mathbb{E}_{{\v{x}\sim p\pr{\v{x}\mid \v{x}_t}}}\br{\argmin_{\hat{\bm{\epsilon}_\theta} \in \mathbb{R}^{n\times d}} \norm{\frac{1}{\alpha_t}\v{x}_t- \frac{\sigma_t}{\alpha_t}\hat{\bm{\epsilon}}_\theta  - \v{x}}^2}\\
&\hat{\bm{\epsilon}}_{\text{MSE}}^* = \frac{1}{\sigma_t} \mathbb{E}_{{\v{x}\sim p\pr{\v{x}\mid \v{x}_t}}}\br{{\v{x}_t - \alpha_t \v{x}}}
\end{align}
Using Tweedie's formula, we obtain the usual score matching relationship:
\begin{align}
&\hat{\bm{\epsilon}}_{\text{MSE}}^* = -{\sigma_t} \nabla_{\v{x}_t} \log p\pr{\v{x}_t}.\label{eq:estimator}
\end{align}

To prove our result, we consider the energy loss objective,
\begin{align}
\label{eq:energy_diffusion}
\mathcal{J}_t\pr{\v{x}_t, \theta} = \mathbb{E}_{{\v{x}\sim p(\v{x}\mid \v{x}_t)}} \br{ E\pr{\frac{1}{\alpha_t}\v{x}_t- \frac{\sigma_t}{\alpha_t}\hat{\bm{\epsilon}}_\theta , \v{x}}}
\end{align}
where
\begin{align}
E(\hat{\v{x}}, \v{x}) &= \sum_{i,j}^n  \left(\lVert\v{x}_i -\v{x}_j \rVert - \lVert\hat{\v{x}}_i -\hat{\v{x}}_j \rVert \right)^2 = \sum_{i,j}^n  \left( d\pr{\v{x}}_{ij} - d\pr{\hat{\v{x}}}_{ij} \right)^2
\end{align}
and $d: \mathbb{R}^{n\times d} \to \mathbb{R}^{n^2}$ computes the distance matrix. 

We perform a Taylor approximation of $d\pr{{\v{x}}}$ and $d\pr{\hat{\v{x}}}$ around $\v{x}_t$
\begin{align}
d\pr{{\v{x}}} \approx d\pr{\v{x}_t} + J\pr{\v{x}_t} \pr{\v{x} - \v{x}_t}
\end{align}
where $J\pr{\hat{\v{y}}}$ is the Jacobian of $d$.

This approximation is valid when $\abs{\v{x} - \v{x}_t}\ll 1$ and $\abs{\v{x} - \hat{\v{x}}_t}\ll 1$ which corresponds to $\abs{\sigma_t} \ll \frac{1}{\bm{\epsilon}}$ and $\abs{\sigma_t} \ll \frac{1}{\hat{\bm{\epsilon}}}$ respectively. Assuming that $\bm{\epsilon}$ follows a standard normal distribution, and that $\hat{{\bm\epsilon}}$ is bounded, for usual diffusion schedules (with monotonically increasing noise) the approximation will be valid for small values of $t$.

This yields
\begin{align}
E(\hat{\v{x}}, \v{x}) &= \norm{ J\pr{\v{x}_t} \pr{\v{x} - \hat{\v{x}}} }^2
\end{align}
Replacing in \cref{eq:energy_diffusion}, we have
\begin{align}
\label{eq:approx_energy}
\mathcal{J}_t\pr{\v{x}_t, \theta} = \mathbb{E}_{{\v{x}\sim p(\v{x}\mid \v{x}_t)}} \br{\norm{ J\pr{\v{x}_t} \pr{\frac{1}{\alpha_t}\v{x}_t - \frac{\sigma_t}{\alpha_t}\hat{\bm{\epsilon}}_\theta - \v{x}} }^2}
\end{align}
Since $\mathcal{J}_t$ is a convex function of $\hat{\bm{\epsilon}}$, its minimization with respect to $\hat{\bm{\epsilon}}_\theta$ can be performed by solving for vanishing gradient, which leads to the minimizer $\hat{\bm{\epsilon}}^*$ satisfying
\begin{align}
&J\pr{\v{x}_t}^TJ\pr{{\v{x}}_t} \pr{\hat{\bm{\epsilon}}^* - \frac{1}{\sigma_t}\mathbb{E}_{{\v{x}\sim p(\v{x}\mid \v{x}_t)}} \br{ \v{x}_t - \alpha_t\v{x}}} = 0  \\
&J\pr{\v{x}_t}^TJ\pr{\v{x}_t} \pr{\hat{\bm{\epsilon}}^* - \sigma_t \nabla_{\v{x}_t} \log p\pr{\v{x}_t}} = 0  
\end{align}
Because of the $SE(d)$ invariance of $d$, $J\pr{\v{x}_t}$ does not have full-rank. Denoting the kernel space of $J\pr{\v{x}_t}$ as $\mathrm{ker}\pr{J\pr{\v{x}_t} }$, we obtain
\begin{align}
\hat{\bm{\epsilon}}^* = - \v{P}_{J\pr{\v{x}_t}} \frac{1}{\sigma_t} \nabla_{\v{x}_t} \log p\pr{\v{x}_t} + \v{v}, \quad  \v{v} \in \mathrm{ker}\pr{J\pr{\v{x}_t}}
\end{align}
where $\v{P}_{J\pr{\v{x}_t}}$ is the orthogonal projector onto $\mathrm{range}\pr{J\pr{\v{x}_t}}$. 

Using the assumption that the measure $p\pr{\v{x}_t}$ is $SE(d)$-invariant, the score lies in $\mathrm{range}\pr{J\pr{\v{x}_t}}$ and is orthogonal to any $\v{v} \in \mathrm{ker}\pr{J\pr{\v{x}_t}}$. The norm of $\hat{\bm{\epsilon}}^* $ is therefore minimized when it is equal to the score. This completes the proof.

\end{proof}

\subsubsection{Proof of \cref{prop:variance}}

\begin{proof}
We assume the same approximation regime as in \cref{prop:score}.

The Monte-Carlo estimators associated with the two losses (for the distance loss, we assume the minimum norm minimizer) are given by
\begin{align}
& \hat{\bm{\epsilon}}_{\text{dist}}^* = - \v{P}_{J\pr{\v{x}_t}} \frac{1}{\sigma_t} \frac{1}{N} \sum_i^N \v{x}_t - \alpha_t\v{x}^{(i)} && \hat{\bm{\epsilon}}_{\text{MSE}}^* = - \frac{1}{\sigma_t} \frac{1}{N} \sum_i^N \v{x}_t - \alpha_t\v{x}^{(i)} 
\end{align}
where the $N$ samples $\v{x}^{(i)}$ are drawn i.i.d. from $p(\v{x}\mid\v{x}_t)$.

$\text{Bias}\br{\hat{\bm{\epsilon}}^*_{\text{dist}}} = 0$:  

This simply follows from \cref{eq:estimator} and the fact that the true score for an $SE(d)$-invariant density lies in $\mathrm{range}\pr{J\pr{\v{x}_t}}$, so that $- \frac{1}{\sigma_t} \v{P}_{J\pr{\v{x}_t}} \nabla_{\v{x}_t} p\pr{\v{x}_t} = - \frac{1}{\sigma_t} \nabla_{\v{x}_t} p\pr{\v{x}_t}$.

$\text{Var}\br{\hat{\bm{\epsilon}}^*_{\text{dist}}} \lesssim \text{Var}\br{\hat{\bm{\epsilon}}^*_{\text{MSE}}}$:

We have
\begin{align}
\text{Cov}\br{\hat{\bm{\epsilon}}^*_{\text{dist}}} = \v{P}_{J\pr{\v{x}_t}}\text{Cov}\br{\hat{\bm{\epsilon}}^*_{\text{MSE}}}\v{P}^T_{J\pr{\v{x}_t}}
\end{align}

We can then obtain,
\begin{align}
&\mathrm{Tr}\pr{\text{Cov}\br{\hat{\bm{\epsilon}}^*_{\text{dist}}}} = \mathrm{Tr}\pr{{\v{P}_{J\pr{\v{x}_t}}}\text{Cov}\br{\hat{\bm{\epsilon}}^*_{\text{MSE}}}{\v{P}^T_{J\pr{\v{x}_t}}}}\\
&\mathrm{Tr}\pr{\text{Cov}\br{\hat{\bm{\epsilon}}^*_{\text{dist}}}} = \mathrm{Tr}\pr{{\v{P}_{J\pr{\v{x}_t}}}\text{Cov}\br{\hat{\bm{\epsilon}}^*_{\text{MSE}}}}
\end{align}
which compared to
\begin{align}
&\mathrm{Tr}\pr{\text{Cov}\br{\hat{\bm{\epsilon}}^*_{\text{MSE}}}} =  \mathrm{Tr}\pr{{\v{P}_{J\pr{\v{x}_t}}}\text{Cov}\br{\hat{\bm{\epsilon}}^*_{\text{MSE}}}}
+ \mathrm{Tr}\pr{\pr{\v{I} - {\v{P}_{J\pr{\v{x}_t}}}}\text{Cov}\br{\hat{\bm{\epsilon}}^*_{\text{MSE}}}}
\end{align}
since $\pr{\v{I} - {\v{P}_{J\pr{\v{x}_t}}}}$ and ${\v{P}_{J\pr{\v{x}_t}}}$ are both positive semi-definite, we conclude
\begin{align}
\mathrm{Tr}\pr{\text{Cov}\br{\hat{\bm{\epsilon}}^*_{\text{dist}}}} \leq \mathrm{Tr}\pr{\text{Cov}\br{\hat{\bm{\epsilon}}^*_{\text{MSE}}}} 
\end{align}
which completes the proof.
\end{proof}

At a high-level, the projector ${\v{P}_{J\pr{\v{x}_t}}}$ removes the components associated with rigid motions from the estimated score. The minimum norm assumption amounts to the network denoinsing to the closet sample to $\v{x}_t$ in the orbit of $\v{x}$.

\subsection{The Boltzmann Distribution is the Steady-state Distribution}
\label{apd:boltzmann}

We consider the potential energy $E(\hat{\v{y}}, \v{y})$ around a configuration $\v{y}$. We assume $\v{y}$ in an approximate equilibrium configuration of the system, and therefore a local minimum of the potential. We model the evolution of the system using Newton's equation (in Hamiltonian form), with an additional term modeling the noise. This gives the SDEs
\begin{align}
    & \diff \hat{\v{y}} = \pd{K}{\v{P}} \diff t, 
    && \diff {\v{P}} = -\pd{E}{\hat{\v{y}}} \diff t -  \v{P} \diff t + \sqrt{2 T} \diff \v{W}.
\end{align}
where $\v{W}$ is generically taken as standard Brownian noise and $K$ is the kinetic energy of the system. The term proportional to $\v{P}$ in the second equation represents the effect of friction and has the effect of bringing the system back towards the equilibrium configuration. This is the Langevin equation, introduced by \cite{einstein1905motion, langevin1908theory} to study Brownian motion. It describes the evolution of a system in a heat bath at temperature $T$; this is the uncertainty model we consider. 

We suppose momentum variables are not of interest. We therefore take the kinetic energy (or masses) as negligible, obtaining the \textit{overdamped} limit \cite{kramers1940brownian}
\begin{align}
\diff {\v{y}} = -\pd{E}{\hat{\v{y}}} \diff t + \sqrt{2 T} \diff \v{W}
\end{align}
The probability density of $\v{y}$ evolves according to the Fokker-Plank equation \cite{chandrasekhar1943stochastic}
\begin{align}
\pd{p(\hat{\v{y}}, t\mid \v{y})}{t} = T \nabla_{\v{y}}^2 {p(\hat{\v{y}}, t\mid \v{y})} + \nabla_{\v{y}} \cdot \br{p(\hat{\v{y}}, t\mid \v{y})) \nabla_{\v{y}} E(\hat{\v{y}},\v{y})}
\end{align}
For anything but the simplest potentials, this equation admits no known closed-form solution. However, under some growth conditions on the potential $V(\v{y})$ (see e.g. \cite{gardiner1985handbook}), the Fokker-Plank equation admits the Boltzmann distribution as a unique steady-state solution when $t\to \infty$:
\begin{align}\label{eq:fokker}
p(\hat{\v{y}}\mid \v{y}) = \frac{1}{Z(\v{y}, T)}\exp\pr{- E(\hat{\v{y}}, \v{y}) / T}
\end{align}
where $Z(\v{y}, T)$ is the partition function.

\subsection{Second-order Taylor Approximations of Potential Energies}
\label{apd:taylor}

Below we justify the two practical choices for the coefficients $k_{ij}\pr{\v{y}}$ (constant and inverse--squared decay) by performing second--order Taylor expansions of two standard pair potentials around their equilibrium bond length $r$. Throughout we write the equilibrium distance between atoms $r=\lVert\mathbf y_1-\mathbf y_2\rVert$ and the difference between the equilibrium distance and the prediction $\delta r=\hat{r}-r$.

\paragraph{Morse potential}
The Morse potential (typically used to describe bonded pairs) is given
\begin{equation}
E_{\text{Morse}}(\hat{r}, r)=D\Bigl[1-\exp\bigl(-a(\hat{r}-r)\bigr)\Bigr]^{2}.\tag{A.2}
\end{equation}
Expanding the exponential for small $\delta r$ gives $\exp(-ax)=1-ax+\tfrac12a^{2}\delta r^{2}+\mathcal O(\delta r^{3})$.  The quadratic approximation is
\begin{align}
E_{\text{Morse}}(\hat{r}, r) &= Da^{2}\delta r^{2} \
= \tfrac12,k^{(\text{M})}\delta r^{2}
\end{align}
with
\begin{equation}
{k^{(\text{M})}=2Da^{2}}\tag{A.4}
\end{equation}
Because $D$ and $a$ are fixed per bond type, $k^{(\text{M})}$ is constant in $r$.

\paragraph{Lennard--Jones potential}
The 12--6 Lennard--Jones potential (commonly used for non-bonding interactions) can be written in terms of its equilibrium separation
\( r=2^{1/6}\sigma\):
\begin{equation}
E_{\text{LJ}}(\hat r,r)
=\varepsilon
\Bigl[
      \bigl(\tfrac{ r}{\hat r}\bigr)^{12}
     -2\bigl(\tfrac{ r}{\hat r}\bigr)^{6}
\Bigr].
\tag{A.5}
\end{equation}
Expanding for small \(\delta r\) and keeping terms up to second order,
\begin{align}
E_{\text{LJ}}(\hat r,r)
&=-\varepsilon
   +\frac12\,k^{(\text{LJ})}\delta r^{2}
   +\mathcal O(\delta r^{3}),\\[4pt]
k^{(\text{LJ})}
&=\left.\frac{\partial^{2}E_{\text{LJ}}}{\partial \hat r^{2}}\right|_{\hat r= r}
 =\frac{72\,\varepsilon}{r^{2}}.
\tag{A.6}
\end{align}
Because \(\varepsilon\) is fixed for a given atom-pair type, the resulting spring constant
\[
{k^{(\text{LJ})}\propto r^{-2}}
\]
decays with the \emph{inverse square} of the equilibrium distance.

\subsection{Invariance and Invariant Loss Functions}
\label{apd:invariant}
Physical systems frequently have inherent symmetry, and machine learning models built for such systems benefit from handling these symmetries.
Two common symmetries are the $E(3)$ symmetry of Cartesian coordinates, and the $S(n)$ symmetry of exchangeable objects (such as atoms in an atomistic systems).
In the context of generative models, we note three approaches to designing models that respect the symmetry of the underlying distribution:
\paragraph{Invariant Distribution}
If we decide that the probability distribution we want to generate is \emph{invariant} to symmetry transformations, a common strategy is to first sample from a distribution that is invariant to the group of interest, and then applying a function that is equivariant to the group \citep{kohler2020equivariant}.
For example, \cite{hoogeboom2022equivariantdiffusionmoleculegeneration} samples from a Gaussian distribution that is invariant to rotations in $SO(3)$, and uses an $E(3)$-equivariant neural network to parametrize a denoising diffusion model.
The probability distribution of the resulting structures generated by their model is therefore invariant to rotations.
\paragraph{Alignment}
In many situations, we want predictions from a neural network to match ground truth data, \emph{up to some symmetry transformation}.
This is often the case in generative models for proteins and molecules, where there is symmetry to SE(3) and S($n$).
Recent works such as \cite{hassan2024flow, klein2023equivariantflowmatching} handle this in a flow-matching context by performing an alignment procedure between samples from a prior and data samples, finding an element of S(n) and SO(3) that minimizes the distance between them.
While costly, this optimal alignment can be found by using the Hungarian algorithm \citep{kurtzberg1962approximation} and the Kabsch algorithm \citep{kabsch1976solution}.
Similarly, \citet{sareen2025symmetryawaregenerativemodelinglearned}
use an equivariant network to learn alignments that brings data samples into canonical representatives \citep{kaba2023equivariance}, which are then fed into a generative model. \cite{zhang2024symdiff} use a similar symmetrization method to achieve equivariance. This technique obviates the need to use an expensive equivariant network for the generative model.

\paragraph{Invariance-based loss}
Another method to handle symmetry is to use a loss based only on invariant features.
Works such as \cite{xu2021learning}, \cite{simm2019generative} and \cite{nesterov20203dmolnet} learn to model interatomic distance matrices rather than atomic coordinates, and convert from distance matrices into coordinates as a post-processing step.
Because distances are invariant under E(3) transformations, the resulting method is invariant.
Our proposed loss function \cref{eq:ploss} falls  into this last category.

\subsection{More on Rigidity Theory}
\label{apd:rigidity}
\paragraph{Background on theory}
We begin by describing the setting of rigidity theory and the necessary properties for our energy loss to scale linearly in the number of vertices.

Rigidity theory defines a \textit{framework} as a graph $G=(V,E)$ and a map $\phi: V \rightarrow \mathbb{R}^d$ which can be interpreted as the physical coordinates of a given vertex. We call a framework \textit{globally rigid} if every $\phi': V \rightarrow \mathbb{R}^d$ that yields the same distances between adjacent vertices is obtained from $\phi$ by an isometry. A framework is \textit{rigid} if there are no \textit{non-trivial} continuous motions of vertices starting from $\phi$ that preserves the distance between adjacent vertices. Trivial motions, in this case, correspond to group elements of $E(d)$.
The central problem of rigidity theory is to determine under which conditions different families of frameworks are rigid or globally rigid \citep{rigidity_talk} \citep{thorpe1999rigidity}.

In our application, it suffices to neglect modeling some interactions between atoms (terms in the sum in \cref{eq:loss}), as long as the minimum loss configuration is unique and thus corresponds to the data. In other words, we require the edges describing this sum are \textit{globally rigid}.

Due to the simplicity of their construction, we mention a few recent results on the rigidity of random graphs. The first result is that a random $k$-regular graph is rigid with high probability (w.h.p.) in $D$ dimensions for $k \geq D^2$. 
It has been conjectured that $k \geq 2D$ should be enough for rigidity, but the existing proof is limited to $D=2$ \citep{krivelevich2023rigid} \citep{rigidity_talk}.
Alternatively, if using Erdos-Renyi random graphs, in $D$ dimensions, one could keep adding edges at random until the minimum node degree becomes $D$, at which point the graph becomes rigid w.h.p. \citep{blms.12740}. Note that a promising construction for rigid graphs was also recently proposed in the context of machine learning \citep{wang2025a}.

\paragraph{2$D$-regular graphs are globally rigid w.h.p.}
To construct a sparse version of the energy loss, we require that the edges over which the pairwise distances differences are summed in \cref{eq:loss} make up a globally rigid graph. For simplicity, we use random 2$D$-regular graphs in the sparse version of the energy loss since the number of edges scales linearly in $N$.

\begin{figure}
    \centering
    \includegraphics[width=0.6\linewidth]{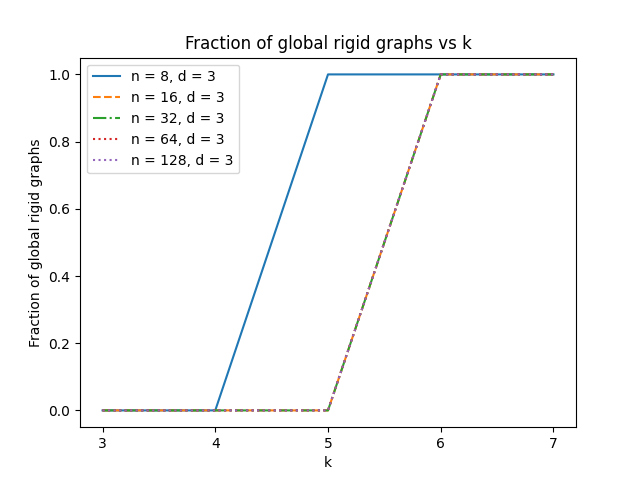}
    \caption{Global rigidity testing of random $k$-regular graphs. Here, $n$ denotes the number of vertices and $d$ the dimension.}
    \label{fig:rigidity}
\end{figure}

We empirically confirm the conjecture for $D$ and $N$ in ranges relevant for our setting. We construct 1000 random 6-regular graphs and check the fraction of the graphs that are globally rigid using rigidity checking code from \citet{dewar2024rigidity}. This code implements a rank check of the rigidity matrix and a random stress test.
The key result is depicted in \cref{fig:rigidity}.

These graphs are used in the sparse energy loss for shape prediction. For molecules, we use a symmetrized version of these graphs according to molecule symmetries. In both settings, 100 random graphs are pre-generated for each number of vertices and a random graph from this pool is chosen when computing the loss.

\paragraph{Symmetrization procedure for random graphs}
We notice that \cref{prop:symmetry} does not necessarily hold for \cref{eq:loss} when the edges are not complete. This means atoms symmetric under $\mathrm{Aut}\pr{\Delta y}$ will not necessarily obtain the same gradients. To remedy this, we introduce a symmetrization procedure for a given $k$-regular graph and molecule.

We can select $k$ random edges for a representative node in each orbit and symmetrize the adjacency matrix over the orbits via

\begin{align}
    A_{sym} = \sum_{g\in \mathrm{Aut}\pr{\Delta y}} g \cdot A \cdot g^{-1}.
\end{align}

Notice that the total number of edges we get by this procedure is

\begin{align}
    |E_{sym}| = O\pr{\sum_{uv \in E} |uG||vG|},
\end{align}

which can be quadratic in $|V|$ if there are large orbits (ie. size $O(|V|)$) directly connected by edges but is linear otherwise. It may be possible to ensure this is linear w.h.p. by choosing the $k$ edges in a way that depends on $G$ but we leave this for future work. We empirically verify these edges are sparse in our setting. 

\subsection{More on Spin Energy}
\label{apd:spin}

For an Ising configuration $\mathbf y\in\{-1,+1\}^{\Lambda}$ the energy change caused by flipping spin $y_i$ is $\Delta E_i = 2 \v{y}_i h_i^{\text{LF}}(\mathbf y)$ with $h_i^{\text{LF}}(\mathbf y)=\sum_j J_{ij}\v{y}_j$.
Setting $h_0=0$ therefore makes
\[
E_{\text{LF}}(\hat{\mathbf y},\mathbf y)
   =\sum_{i\in\Lambda} h_i^{\text{LF}}(\mathbf y)\,\hat{\v{y}}_i,
\]
proportional to this exact spin-flip energy around the ground state. The local energy we propose is therefore a sensible linear approximation to the true energy.

To obtain a convex loss, we make sure the local field weighting is always positive, For this, we add a global offset $h_0>0$ in $h_i^{\text{LF}}(\mathbf y)=\sum_j (J_{ij} + h_0)\v{y}_j$.
Since each site has at most four neighbours with $|J_{ij}|\le1$, we have $-4\le h_i^{\text{true}}\le4$; choosing a single $h_0>4$ ensures weight $(J_{ij} + h_0)>0$ and penalises energetically costly errors more heavily.

\subsection{Extension to Flow Matching}
The energy loss can be extended to Gaussian flow matching \citep{lipman2023flow}. We show hereafter the correspondence for the conditional vector field. The noisy sample is given by the interpolation $$ \v{x}_t = (1-t) \v{x} + t \bm{\epsilon} .$$

The flow matching objective aims at regressing the vector field: $$ \v{u} = \frac{\v{x}_t - \v{x}}{t} .$$

Given a vector field prediction, the corresponding sample prediction is $$ \v{x} = \v{x}_t - t \v{u}_\theta(\v{x}_t) .$$

The correspondence between MSE on the vector field prediction and on sample prediction is therefore: $$ \lVert \v{u} - \v{u}_\theta \rVert^2 = \frac{1}{t^2} \lVert \v{x} - \v{x}_\theta \rVert^2 .$$

Therefore, the associated energy objective is obtained by replacement of the regression MSE:
$$ \frac{1}{t^2} E (\v{x} - \v{x}_\theta) .$$

Our theoretical results relating to score estimation properties also transfer to flow matching. This is because Gaussian flow matching also implicitly provides a method for score estimation similar to diffusion models. Given the optimal vector field, $\v{u}^*(\v{x}_t)$ , the score is given by:
$$ \nabla_{\v{x}_t} \log p(\v{x}_t) = -\frac{(1-t) \v{u}^*(\v{x}_t) + \v{x}_t}{t}. $$

\section{Related works}

Several different lines of research have also incorporated a concept of energy into a machine learning framework. In this section, we distinguish our framework from distinct but related areas of research.

\paragraph{Energy-based models} Traditional energy-based models approach learning as shaping an energy landscape, where observed configurations correspond to low-energy states \citep{lecun2006tutorial}. Deep counterparts and their connection to discriminative training have also been extensively explored in many recent works (e.g., \cite{du2019implicit,grathwohl2019your}).
A key distinction of the existing literature on energy-based models and our energy loss approach is that, because they minimize the forward KL (max-likelihood or alternatives such as contrastive and large-margin losses), they need to deal with the minimization of the partition function or its surrogates -- i.e., the energy of arbitrary points in the domain must be high. In contrast, our treatment remains close to supervised learning losses and avoids the partition function altogether.

\paragraph{Physics-informed neural networks (PINNs)} \citep{raissi2019physics} has proposed PINNs as a way to learn PDEs by penalizing residuals directly in the loss, enforcing solutions consistent with physical constraints. They have recently been used in the context of diffusion models \citep{bastek2024physics}. In contrast to our approach, these models do not rely on training data and models are instead learned to satisfy known differential equations on randomly generated points from the domain. 
Another family of physics-informed losses appears in Hamiltonian Neural Networks (HNN) \citep{greydanus2019hamiltonian} and Lagrangian Neural Networks (LNN) \citep{cranmer2020lagrangian}.

\paragraph{Energy Sampling and Boltzmann Generators}
A separate line of research incorporating energies and generative modeling has been in Boltzmann Generators \citep{noe2019boltzmann, kohler2020equivariant, klein2025transferableboltzmanngenerators}. These models are designed to sample physical configurations according to a Boltzmann distribution stemming from a known energy function. 
While our framework is also based on an assumption of data belonging to a Boltzmann distribution, ours is instead simply an approximation of the local landscape around each data point and does not assume the existence of a callable energy function.

\section{Experimental Details and Additional Results}
\label{apd:experiments}
\subsection{Regular shape generation}\label{apd:shape}
\paragraph{Experimental details and hyperparameters}
We use a 2 hidden-layer MLP with hidden dimension 64 for this task. We conducted a sweep over hidden dimension and find behaviour is relatively consistent. Models are trained in parallel on an Nvidia Quadro RTX 8000 using the Adam optimizer. The dataset size is 100K randomly generated samples and we train all models for 50 epochs. A sweep over dataset size showed fairly consistent results. We conduct thorough sweeps for learning rate for each loss, shape degree and augmentation angle. For each setting, the model giving the highest quality is chosen.

\paragraph{Evaluation}
To evaluate shape quality, we introduce a metric that captures how regular the angular differences and radial distances are across the shape.
In a well-formed, regular shape, we expect both the variation in angular differences ($\sigma_{\Delta_{angle}}$) and the variation in radial distances ($\sigma_{radius}$) to be small.
In particular, we choose $\text{Quality} \coloneq -\ln (\frac{\sigma_{\Delta_{angle}}}{2\pi} + \frac{\sigma_{radius}}r)$. This is, of course, a design choice used to map a shape to a single number. We record both $\sigma_{\Delta_{angle}}$ and $\sigma_{radius}$ to ensure both terms are well-represented in the quality and find that visually this metric is a good reflection of the visual regularity of a generated shape. For reference, above a quality of 5-6, shapes look nearly visually perfect, as in \cref{fig:shape_generation}a. For quality below this, they become slightly irregular and below 2 they look very disordered as in \cref{fig:shape_generation}b.

\subsection{Molecule generation}\label{apd:molecule}
\begin{table}[t]
\vspace{-4ex}
\centering
\caption{\footnotesize{Best hyperparameters for GDM.}}
  \vspace{1em}
\label{tab:gdm_hparams}
\begin{tabular}{llcc}
\toprule
\textbf{Loss} & \textbf{Coefficient} & \textbf{Learning Rate} & \textbf{Positional Loss Weight} \\
\midrule
Energy
& Constant   & 9e-4   & 0.05 \\
& Inv. Dist.  & 7e-4   & 0.05 \\
& Inv. Sq. Dist. & 6e-4   & 0.05 \\
& Exp. Dist.  & 4e-4 & 0.05 \\
\midrule
MSE              & - & 1e-3  & 1.5 \\
MAE              & - & 1e-3  & 0.8 \\
Kabsch Align     & - & 8e-4  & 0.8 \\
\bottomrule
\end{tabular}
\end{table}

\begin{table}[t]
\centering
\caption{\footnotesize{Best hyperparameters for EDM.}}
\vspace{1em}
\label{tab:edm_hparams}
\begin{tabular}{llcc}
\toprule
\textbf{Loss} & \textbf{Coefficient} & \textbf{Learning Rate} & \textbf{Positional Loss Weight} \\
\midrule
Energy
& Constant        & 1e-4 & 0.05 \\
& Inv. Dist.      & 1e-4 & 0.05 \\
& Inv. Sq. Dist.  & 1e-4 & 0.1 \\
& Exp. Dist.      & 1e-4 & 0.05 \\
\midrule
MSE                  & - & 3e-4 & 1.0 \\
MAE                  & - & 3e-4 & 1.0 \\
Kabsch Align         & - & 3e-4 & 0.8 \\
\bottomrule
\end{tabular}
\end{table}

\subsubsection{QM9 Dataset}
\paragraph{Experimental details and hyperparameters}
On QM9, we match the setup in \citet{hoogeboom2022equivariantdiffusionmoleculegeneration} as closely as possible. We train GDMs and EDMs with with 9 layers and 256 node features on 100k samples from the dataset. The diffusion process has 1000 diffusion steps
with polynomial noise schedule and precision \num{1e-5}. An L2 denoising loss is used with mini-batch size 512 on GDM and 400 on EDM. We use the Adam optimizer. An EMA decay of 0.9999 is used. Runs were conducted on single 48G GPUs mainly on the Nvidia Quadro RTX 8000, A6000 and L40S. A full run of 3000 epochs takes 2-4 days on a single GPU.

We conduct extensive sweeps for learning rate and positional loss weight for all losses. We tune the positional loss weight to ensure there is balance between loss on positions and atom-type for all losses. Learning rates were searched for in broadly in the range [1e-5, 1e-2] before narrowing the range to [2e-3, 4e-4]. For the positional loss weight, we choose values in $[0.05, 0.1, 0.5, 0.8, 1.0, 1.5]$. We find final performance is not very sensitive to the positional loss weight. Tuned hyperparameters are summarized in \cref{tab:gdm_hparams} and \cref{tab:edm_hparams}.

\paragraph{Additional results}
Here, we include results for all settings for GDM, GDM-aug and EDM. The results follow in \cref{tab:all_results}. Results are averaged across seeds.

\begin{table}[t]
\vspace{-4ex}
  \caption{\footnotesize{Complete results on QM9.}}
  \vspace{1em}
  \label{tab:all_results}
  \centering
  \begin{tabular}{lcccc}
    \toprule
    Loss & Mol. stab. (\%) & Atom stab. (\%) & Valid. (\%) & Unique (\%) \\
    \midrule
    \multicolumn{5}{l}{\textbf{GDM}} \\
    \midrule
MSE & 81.7 $\pm$ 3.3 & 98.3 $\pm$ 0.3 & 93.3 $\pm$ 1.7 & 99.98 $\pm$ 0.04 \\
MAE & 76.3 $\pm$ 2.0 & 97.7 $\pm$ 0.3 & 91.1 $\pm$ 1.2 & 99.96 $\pm$ 0.05 \\
Kabsch Align & 81.7 $\pm$ 2.2 & 98.4 $\pm$ 0.2 & 93.1 $\pm$ 1.2 & 99.93 $\pm$ 0.13 \\
Energy (Sparse) & \textbf{86.1} $\pm$ 2.3 & \textbf{99.0} $\pm$ 0.1 & 96.2 $\pm$ 1.4 & 100.0 $\pm$ 0.0 \\
Energy & \textbf{86.2} $\pm$ 2.1 & 98.9 $\pm$ 0.2 & \textbf{96.6} $\pm$ 1.3 & 100.0 $\pm$ 0.0 \\
    \midrule
    \multicolumn{5}{l}{\textbf{GDM-aug}} \\
    \midrule
MSE & 83.7 $\pm$ 2.3 & 98.3 $\pm$ 0.004 & 93.6 $\pm$ 1.7 & 100.0 $\pm$ 0.0 \\
MAE & 76.4 $\pm$ 0.9 & 98.1 $\pm$ 0.3 & 92.6 $\pm$ 1.2 & 99.99 $\pm$ 0.02 \\
Kabsch Align & 82.3 $\pm$ 0.5 & 97.8 $\pm$ 0.004 & 90.8 $\pm$ 2.0 & 100.0 $\pm$ 0.0 \\
Energy (Sparse) & 89.1 $\pm$ 0.9 & 99.0 $\pm$ 0.1 & 97.4 $\pm$ 2.5 & 100.0 $\pm$ 0.0 \\
Energy & \textbf{89.8} $\pm$ 2.8 & \textbf{99.3} $\pm$ 0.3 & \textbf{97.7} $\pm$ 1.4 & 99.99 $\pm$ 0.002 \\
    \midrule
    \multicolumn{5}{l}{\textbf{EDM}} \\
    \midrule
MSE & 82.4 $\pm$ 3.4 & 98.8 $\pm$ 1.7 & 93.0 $\pm$ 2.5 & 99.89 $\pm$ 0.32 \\
MAE & 74.8 $\pm$ 1.7 & 97.8 $\pm$ 0.3 & 88.6 $\pm$ 0.7 & 99.96 $\pm$ 0.07 \\
Kabsch Align & 80.6 $\pm$ 3.0 & 98.3 $\pm$ 3.0 & 92.5 $\pm$ 3.0 & 99.91 $\pm$ 0.07 \\
Energy & \textbf{86.6} $\pm$ 1.6 & \textbf{99.0} $\pm$ 0.20 & \textbf{96.8} $\pm$ 1.1 & 99.96 $\pm$ 0.06 \\\bottomrule
  \end{tabular}
\end{table}

\subsubsection{GEOM-Drugs Dataset}
\paragraph{Experimental details and hyperparameters}
On GEOM-Drugs, we use a similar setting to QM9 but now train models with 4 layers and 256 node features, following \citet{hoogeboom2022equivariantdiffusionmoleculegeneration}. We train the model for 13 epochs. Training is distributed across 4 80G Nvidia A100l GPUs and a single run takes roughly 2.5 days. We use a batch size of 128 with the Adam optimizer.

We start with optimal learning rate and positional loss weight from QM9 and do a sweep over learning rates [5e-4, 1e-3, 2e-3] for MSE and [1e-4, 4e-4, 1e-3] for energy loss.
The hyperparameters in \cref{tab:gdm_hparams} gave the best results. We use exponential coefficients for the energy loss.

\paragraph{Additional results}
We additionally report the performance of MSE and energy losses with GDM-aug in \cref{tab:geom_complete}.

\begin{table}[t]
  \caption{\footnotesize{Complete results for GDM and GDM-aug on GEOM-Drugs.}}
  \vspace{1em}
  \centering
  \begin{tabular}{lcccc}
    \toprule
    Loss & Mol. stab. (\%) & Atom stab. (\%) & Valid. (\%) & Unique (\%) \\
    \midrule
    \multicolumn{5}{l}{\textbf{GDM}} \\
    \midrule
    MSE    & 0.3  & 84.7 & \textbf{93.8} & 100 \\
    Energy & \textbf{21.1} & \textbf{95.8} & 89.6 & 100 \\
    \midrule
    \multicolumn{5}{l}{\textbf{GDM-aug}} \\
    \midrule
    MSE    & 0.8  & 85.6 & \textbf{94.8} & 100 \\
    Energy & \textbf{24.6} & \textbf{96.0} & 89.7 & 100 \\
    \bottomrule
  \end{tabular}
  \label{tab:geom_complete}
\end{table}

\subsection{Spin ground state prediction}
\label{apd:spin}
\paragraph{Experimental details and hyperparameters}

The CNN we use for the spin prediction task is a 6 layer ResNet type architecture with 256 hidden layer size. All networks are trained with a learning rate of $1\times 10^{-3}$ until convergence. We use the Adam optimizer with batch size 256. Temperature is set to $T = 0.1$ for the local energy loss. Training takes around 5 hours on Nvidia V100 GPUs.

\subsection{More on sparse energy loss}
\label{apd:sparse_timing}
\subsubsection{Timing}
Our objective in including the sparse energy loss is to demonstrate our method can efficiently generalize to systems with many particles where loss calculation may contribute significantly to running time (e.g. very large point clouds). This is not the case for molecules, where the neural network (GNN or Transformer) is typically fully connected and thus scales as $N^2$. As \autoref{tab:timing_results} shows, the most expensive loss calculation is less than 1\% of the total backward and forward time.

\begin{table}[h!]
\vspace{-2ex}
\centering
\caption{\footnotesize Wall-times for loss computation on QM9 on an NVIDIA L40S.}
\begin{tabular}{l l c}
\toprule
\textbf{Component} & \textbf{Loss Type} & \textbf{Time (ms)} \\
\midrule
\textbf{Loss computation} & MSE            & 0.18 $\pm$ 0.01 \\
                          & Energy         & 0.51 $\pm$ 0.02 \\
                          & Sparse Energy  & 0.57 $\pm$ 0.02 \\
                          & Kabsch Align   & 1.14 $\pm$ 0.03 \\
\textbf{Forward pass}     & –              & 74 $\pm$ 16 \\
\textbf{Backward pass}    & –              & 94 $\pm$ 3 \\
\textbf{Optimizer step}   & –              & 1.43 $\pm$ 0.01 \\
\bottomrule
\end{tabular}
\label{tab:timing_results}
\end{table}

To better understand the scale at which this becomes a relevant consideration and the utility of the sparse energy loss, see the following wall clock times from the shape generation setting in \autoref{tab:node_scaling}.

\begin{table}[h!]
\vspace{-2ex}
\centering
\caption{\footnotesize Runtime for different loss functions as the number of nodes increases.}
\begin{tabular}{c c c c c}
\toprule
\textbf{\# Nodes} & \textbf{Energy (ms)} & \textbf{Sparse Energy (ms)} & \textbf{MSE (ms)} & \textbf{Kabsch Align (ms)} \\
\midrule
30      & 0.240 $\pm$ 0.0021  & 0.255 $\pm$ 0.0012  & 0.058 $\pm$ 0.0004  & 0.807 $\pm$ 0.0027 \\
300     & 0.245 $\pm$ 0.0011  & 0.257 $\pm$ 0.0024  & 0.059 $\pm$ 0.0004  & 0.804 $\pm$ 0.0048 \\
3000    & 20.120 $\pm$ 0.0055 & 0.275 $\pm$ 0.0018  & 0.0779 $\pm$ 0.0008 & 0.944 $\pm$ 0.0028 \\
30000   & --                   & 0.293 $\pm$ 0.0029  & 0.0740 $\pm$ 0.0011 & 3.242 $\pm$ 0.1452 \\
300000  & --                   & 2.652 $\pm$ 0.0050  & 0.131 $\pm$ 0.0004  & 24.381 $\pm$ 0.0272 \\
\bottomrule
\end{tabular}
\label{tab:node_scaling}
\end{table}

At 30000+ nodes, the energy loss requires too much memory to compute. Importantly, the sparse energy is cheaper than the Kabsch Align by a factor 5-10x. Note that all losses have some constant cost that does not scale with N contributing to the wall-clock time. This explains why for QM9 (avg. 28 atoms) energy loss is marginally faster than sparse energy and why in the scaling table wall-clock times start to increase with N only after a certain point.

Interestingly, using an equivariant network with EDM takes 129.71 ± 0.044 ms for the forward pass and 180.07 ± 0.070 ms for the backward pass. Using the energy loss imparts a 0.3\% increase on one backward pass through the model, while using an equivariant architecture imparts a 94\% increase, while providing inferior benefits. The energy loss with a non-equivariant architecture results in more improvement than using an equivariant architecture, at negligible computational cost, which we think is a significant finding.

\subsection{Sparse energy loss on larger systems}

\begin{table}[h!]
\vspace{-2ex}
\centering
\caption{\footnotesize Sparse energy loss with GDM on GEOM-Drugs.}
\begin{tabular}{l c c c c}
\toprule
\textbf{Loss} & \textbf{Mol. stab. (\%)} & \textbf{Atom stab. (\%)} & \textbf{Valid. (\%)} & \textbf{Unique (\%)} \\
\midrule
MSE                       & 0.3  & 84.7  & 93.8  & 100 \\
Energy                    & \textbf{21.1} & \textbf{95.8} & 89.6  & 100 \\
Sparse Energy (Inv. Dist) & 7.4  & 91.9  & 92.6  & 100 \\
\bottomrule
\end{tabular}
\label{tab:sparse_results}
\end{table}

We include the sparse energy results on Geom-Drugs in \autoref{tab:sparse_results}. We found using a more gradual distance decay in the coefficient worked better when the edges are sparse and random. These results highlight a potential compute-performance tradeoff for this version of the loss on larger graphs. This result can likely be improved by being more intentional in the selection of sparse edges. We leave this to future work.

\end{document}